\documentclass{zgroup/zgroup}
\usepackage[utf8]{inputenc}

\usepackage{times}
\usepackage{url}

\usepackage{amssymb}
\usepackage{microtype}
\usepackage{appendix}
\usepackage{caption}
\usepackage{graphicx, color}
\usepackage{subcaption}
\usepackage{booktabs}       
\usepackage{amsfonts}       
\usepackage{nicefrac}       
\usepackage{microtype}      
\usepackage{multirow, multicol}
\usepackage{ragged2e}
\usepackage{amsmath}
\usepackage{bm, bbm}
\usepackage{booktabs} 
\usepackage{enumitem}
\usepackage{tikz}
\usepackage{wrapfig}
\usetikzlibrary{positioning}
\usepackage{lipsum}
\usepackage{float}
\usepackage{array}
\usepackage{pseudocode}
\usepackage{mathtools}
\usepackage{tcolorbox}

\usepackage{hyperref}
\usepackage{url}
\usepackage{graphicx}
\usepackage{booktabs}

\usepackage{xcolor}
\usepackage{listings}
\usepackage{caption}
\usepackage{newfloat}

\definecolor{lightgray}{gray}{0.97}

\DeclareFloatingEnvironment[
    fileext=lol,
    listname={List of Code Boxes},
    name=Code Box
]{listing}

\lstdefinelanguage{prompt}{
    sensitive=false
}

\lstdefinestyle{promptbox}{
    language=prompt,
    basicstyle=\ttfamily\footnotesize,
    backgroundcolor=\color{lightgray},
    breaklines=true,
    columns=fullflexible,
    keepspaces=true,
    frame=single,
    numbers=none,
    showstringspaces=false,
    xleftmargin=0pt,
    xrightmargin=0pt
}

\usepackage[textsize=tiny]{todonotes}

\title[Binary Agreement for Efficient Unsupervised Confidence Calibration]{Binary Agreement for Efficient Unsupervised Confidence Calibration of Reasoning LLMs}

\author{\Name{Thomas Zollo}
\Email{tpz2105@columbia.edu}\\
\addr Columbia University\\
\Name{Jimmy Wang}
\Email{jw4209@columbia.edu}\\
\addr Columbia University\\
\Name{Richard Zemel}
\Email{zemel@cs.columbia.edu}\\
\addr Columbia University \\
}

\begin{document}

\maketitle

\begin{abstract}
Reasoning language models often lack the calibrated confidence needed for reliable deployment, particularly when correctness labels are unavailable and sampling budgets are limited.
We introduce \emph{Binary Agreement for Self-consistency Calibration} (BASC), which uses self-consistency as an unlabeled proxy for correctness without constructing costly self-consistency targets for each calibration query.
Our key observation is that agreement between an anchor response and a single independent probe is a Bernoulli observation whose conditional mean is the anchor's latent self-consistency.
Pooling these observations across unlabeled queries, BASC supports both recalibrating existing confidence scores and learning more discriminative scores before recalibration, using as few as two responses per calibration query and only one at deployment.
The use of binary isotonic recalibration provides a distribution-free calibration guarantee with respect to latent self-consistency.
Across 5 tasks and 9 reasoning models, BASC substantially improves calibration with respect to correctness.
\end{abstract}

\section{Introduction}

Reliable deployment of reasoning large language models (LLMs) requires confidence estimates that accurately reflect the probability that an answer is correct \citep{bommasani2022opportunitiesrisksfoundationmodels,yang2026reliableresponsiblefoundationmodels}.
This property, known as \emph{confidence calibration} \citep{guo2017calibration, kadavath2022language}, is important for deciding when a model can be trusted to answer, when it should defer, or when additional intervention is needed.
Modern LLMs nevertheless remain poorly calibrated
\citep{leng2024taming, damani2025beyond}, and existing confidence estimation approaches often rely either on correctness labels
\citep{zadrozny2001obtaining, kumar2020verifieduncertaintycalibration, band2024linguistic, damani2025beyond}
or on repeated generations at inference time
\citep{chen2024inside, kuhn2023semantic, duan2024shifting},
both of which may be impractical in resource-constrained settings.

Self-consistency, the rate at which independently sampled responses to a prompt agree on a final answer, offers a useful unlabeled proxy for correctness
\citep{wang2023selfconsistencyimproveschainthought, lyu2025calibrating},
but estimating it typically requires many generations per query.
One approach is to amortize this computation by learning a predictor offline, allowing confidence to be produced from a single response at deployment.
However, constructing accurate self-consistency targets for every calibration query still requires many generations, moving the sampling burden from deployment to offline calibration.
These costs are especially consequential for on-device and edge systems, where memory, latency, and battery are constrained \citep{Wang_2025}; for black-box APIs, where every additional generation incurs monetary cost; and for agentic systems, where confidence may need to be evaluated repeatedly to determine when to act, defer, invoke tools, or escalate \citep{oh-etal-2026-uncertainty}.
Obtaining calibrated confidence therefore remains difficult in these important resource-constrained settings, where labels are scarce and repeated sampling is impractical.

\begin{figure}[!t]
    \centering
    \includegraphics[width=\textwidth]{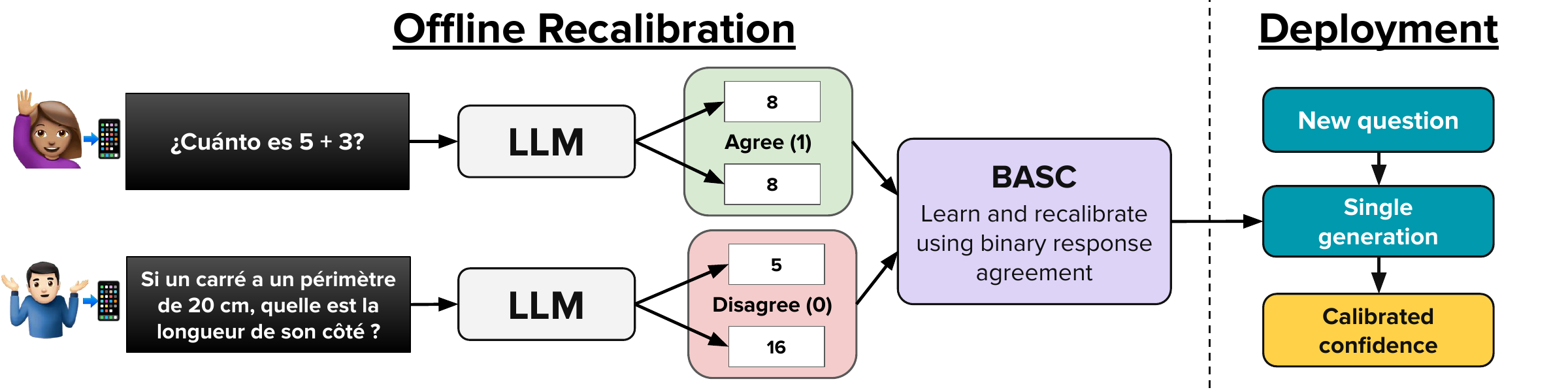}
    \caption{
    Binary Agreement for Self-consistency Calibration (BASC) learns a confidence calibrator offline by pooling binary agreement across unlabeled questions, enabling confidence-guided decisions from a single generation at deployment.
    It targets resource-constrained settings where correctness labels are scarce and latency, memory, or energy budgets make repeated test-time sampling impractical.
    Here, our motivating example is a multilingual on-device math tutor that must decide whether to answer directly or defer to a teacher.
    The figure illustrates $k=2$ responses per question; larger $k$ can be used for score learning.
    }
    \label{fig:basc_overview}
\end{figure}

In this work, we introduce \textbf{\emph{Binary Agreement for Self-consistency Calibration} (BASC)}, an unsupervised framework that reduces sampling costs both offline and at deployment (see Figure \ref{fig:basc_overview}).
Our key observation is that accurately estimating self-consistency for every calibration query is unnecessary.
Instead, we can learn to predict latent self-consistency from binary agreement between just two responses per query.
BASC exploits this signal in two complementary ways.
First, binary agreement can directly recalibrate an existing confidence score, such as token probabilities.
Second, when stronger discrimination is desired or token probabilities are unavailable, these observations can train a score model before recalibration.
In both cases, BASC applies binary isotonic regression to held-out agreement observations and requires only a single response at deployment, 
substantially reducing resource requirements.
Building on \citet{rossellini2026analysis}, this recalibration provides a finite-sample, distribution-free guarantee for calibration to latent self-consistency.

We evaluate BASC across 5 math and question-answering tasks using 9 reasoning models from the Qwen, DeepSeek, and Nemotron families.
Our main contributions and findings are:
\begin{enumerate}
    \item We introduce a binary-agreement framework that avoids costly per-query self-consistency estimation, supporting both label-free recalibration of existing scores and efficient learning of more informative scores.
    \item We show that binary recalibration substantially reduces calibration error with respect to correctness, while agreement-based score learning improves probabilistic quality, discrimination, robustness, and selective prediction.
    \item We justify the recalibration stage with a distribution-free guarantee for single-generation prediction of latent self-consistency, even though this target is never directly observed.
\end{enumerate}
Together, these results establish binary agreement as a practical, inexpensive signal for improving confidence estimates without labels or repeated test-time sampling.

\section{Related Work}
Calibration has been extensively studied for classification \citep{naeini2015obtaining, guo2017calibration, kumar2020verifieduncertaintycalibration, minderer2021revisiting}, but confidence estimation is more difficult for open-ended language generation, where outputs are multi-token and may admit many semantically equivalent forms \citep{kuhn2023semantic, yang2026reliableresponsiblefoundationmodels}.
Common approaches include using token probabilities \citep{kadavath2022language}, verbalized confidence \citep{tian2023just, xiong2024llms, yoon2025reasoning}, and uncertainty estimated from repeated generations, including sample consistency \citep{si2023prompting, lyu2025calibrating} and semantic entropy \citep{kuhn2023semantic}.
Because repeated sampling is expensive, prior work has explored amortizing multi-sample uncertainty signals into single-generation predictors \citep{kossen2024semanticentropyprobesrobust, huang2025efficient}.
These methods reduce inference-time sampling, but treat the uncertainty signal as a per-query target that must be estimated offline from many generations; moreover, \citet{kossen2024semanticentropyprobesrobust} predicts entropy rather than calibrated confidence, while \citet{huang2025efficient} requires LLM tuning and eschews standard self-consistency in favor of a more expensive heuristic target.

Calibration is especially challenging for reasoning models, where post-training tends to induce overconfidence and poor abstention behavior \citep{leng2024taming, kirichenko2025abstentionbench, mei2025reasoning, damani2025beyond}.
Recent approaches address these issues through prompting or introspection \citep{kirichenko2025abstentionbench, mei2025reasoning} or calibration-aware training \citep{band2024linguistic, baniharouni2026rewardingdoubtreinforcementlearning, damani2025beyond}.
These methods generally require additional inference-time elicitation, supervised or reward-based training signals, or modification of the underlying model.
Also related to our unsupervised setting are recent methods that derive confidence from unlabeled signals produced by a paired base model \citep{luo2025pretrainedllmsecretlyunsupervised, tan2026basecalunsupervisedconfidencecalibration}.
However, we find that base-model-derived confidence 
is poorly calibrated in our setting
(see Appendix~\ref{sec:app_base_targets}).

Finally, a separate line of work develops distribution-free methods that provide high-probability guarantees on LLM behavior, often by adapting conformal prediction or related risk-control ideas \citep{zollo2023prompt, quach2023conformal,campos2024conformalpredictionnaturallanguage, dobriban2025statisticalmethodsgenerativeai, ji2025overviewlargelanguagemodels}.
These methods share the broader goal of reliable behavior under uncertainty, but generally do not address answer-level confidence calibration.

\section{Method}\label{sec:method}

We study confidence estimation for reasoning LLMs when correctness labels are unavailable and both offline and inference-time sampling resources are limited.
Let \(x\) denote an input prompt and \(y=(r,a)\sim p_\theta(\cdot\mid x)\) a model response, where \(r\) is the reasoning trace and \(a\) the extracted final answer.
Let \(Z\in\{0,1\}\) indicate whether \(a\) is correct.
Given a single response \(y\) at deployment, our goal is a confidence estimate \(\hat c(x,y)\in[0,1]\) calibrated with respect to correctness:
\[\mathbb{E}[Z\mid C]=C,\] where \(C=\hat c(X,Y)\).
Without correctness labels, we have only an unlabeled calibration set
\(\mathcal{D}_{\mathrm{cal}}=\{x_i\}_{i=1}^n\) drawn i.i.d.\ from the target prompt distribution.

\paragraph{Self-consistency as an unlabeled target.}
Because \(Z\) is unobserved, we use \emph{self-consistency} as an unlabeled proxy for answer confidence
\citep{wang2023selfconsistencyimproveschainthought, lyu2025calibrating}.
For an anchor response \(y=(r,a)\) to prompt \(x\), let
\(Y'=(R',A')\sim\tilde p_\theta(\cdot\mid x)\)
be an independent response drawn from a fixed \emph{probe} sampling distribution \(\tilde p_\theta\).
While the anchor and probe distributions may be the same, separating them allows low-entropy decoding at deployment while using more diverse probe samples to assess uncertainty.

We define the population self-consistency of \(a\) as
\begin{equation}
    s^\star(x,a)
    =
    \mathbb{P}\!\left(A'\equiv a\mid x\right),
    \label{eq:population_sc}
\end{equation}
where \(\equiv\) denotes binary agreement under the task-specific answer-matching procedure
(e.g., exact or numeric matching for math, and semantic matching via bidirectional entailment \citep{kuhn2023semantic} for QA).
Thus, \(s^\star(x,a)\) is the probability that an independent response produces an answer equivalent to \(a\).
Self-consistency is not a correctness probability, but it measures how strongly the model favors its answer and empirically tracks correctness.
Figure~\ref{fig:basc_illustrator}(a) previews this relationship across the 36 model/dataset combinations evaluated in Section~\ref{sec:experiments}, where the gray curve compares empirical self-consistency with answer accuracy.

A standard approach to estimating \(s^\star(x,a)\) uses repeated sampling.
Using \(k\) total responses, draw an \emph{anchor}
\(Y^{(0)}=(R^{(0)},A^{(0)})\sim p_\theta(\cdot\mid x)\)
and \(k-1\) probes
\(Y^{(1)},\ldots,Y^{(k-1)}\) drawn i.i.d.\ from \(\tilde p_\theta(\cdot\mid x)\).
For each probe, define
\(B^{(j)}=\mathbf{1}[A^{(j)}\equiv A^{(0)}]\).
The Monte Carlo estimate of the anchor's self-consistency is
\(\widehat{s}_k=(k-1)^{-1}\sum_{j=1}^{k-1}B^{(j)}\).
Obtaining a low-variance \(\widehat{s}_k\) for every
calibration 
example can require many expensive generations.
Our key observation is that this per-query precision is unnecessary.
Let \(S=s^\star(X,A^{(0)})\).
Conditioned on the prompt and anchor response,
\begin{equation}
    B^{(j)}\mid X,Y^{(0)}
    \overset{\mathrm{i.i.d.}}{\sim}
    \mathrm{Bernoulli}(S),
    \qquad
    \mathbb{E}\!\left[B^{(j)}\mid X,Y^{(0)}\right]=S.
    \label{eq:agreement_expectation}
\end{equation}
Each probe therefore provides an unbiased binary observation of latent self-consistency.
Writing \(B\) for a single probe's agreement and
\(U=u(X,Y^{(0)})\) for any scalar score computed from the anchor by a fixed function \(u\), the tower property gives
\begin{equation}
    \mathbb{E}[B\mid U]
    =
    \mathbb{E}[S\mid U].
    \label{eq:score_agreement_identity}
\end{equation}
Thus, pooling binary agreement observations across queries provides supervision for conditional latent self-consistency without accurately estimating \(S\) for any individual query.

\subsection{Binary Agreement for Self-consistency Calibration (BASC)}

Building on this observation, we propose \textbf{\emph{Binary Agreement for Self-consistency Calibration} (BASC)}, a 2-stage algorithm using binary agreement to learn and recalibrate LLM confidence scores.
Stage~1 supplies a scalar score \(U=u(X,Y^{(0)})\), either pre-existing or learned from agreement observations; Stage~2 recalibrates it on held-out agreement observations.
For a learned score, we split the calibration data into disjoint subsets
\(\mathcal{D}_{\mathrm{cal}}=\mathcal{D}_{\mathrm{train}}\cup\mathcal{D}_{\mathrm{iso}}\)
for score learning and recalibration, respectively.
For a pre-existing score, no training split is needed and
\(\mathcal{D}_{\mathrm{cal}}=\mathcal{D}_{\mathrm{iso}}\).

\paragraph{\emph{Stage 1: Scalar score.}}
The simplest instantiation uses a pre-existing score \(u(x,y)\), such as length-normalized token probability, and requires no score training.
Alternatively, let \(H=h(X,Y^{(0)})\) represent an anchor prompt-response pair
(e.g., the reasoning model's last-token hidden state), and let \(f_\phi(H)\in[0,1]\) predict repeat agreement.
For each \(x_i\in\mathcal{D}_{\mathrm{train}}\), draw one anchor
\(Y_i^{(0)}\sim p_\theta(\cdot\mid x_i)\)
and \(k-1\) probes from \(\tilde p_\theta(\cdot\mid x_i)\).
Set \(H_i=h(x_i,Y_i^{(0)})\) and
\(B_{ij}=\mathbf{1}[A_i^{(j)}\equiv A_i^{(0)}]\), \(j=1,\ldots,k-1\).
For \(k=2\), each anchor contributes one Bernoulli realization, not the latent probability itself; larger \(k\) provides repeated conditionally i.i.d.\ observations for the same anchor.
We train \(f_\phi\) using binary cross-entropy over these outcomes.
This Bernoulli log-loss has Bayes-optimal prediction
\(\mathbb{P}(B=1\mid H)=\mathbb{E}[S\mid H]\),
by the argument of Equation~\ref{eq:score_agreement_identity}, making it a natural objective for predicting conditional latent self-consistency.
The learned score is \(u(x,y)=f_\phi(h(x,y))\).

\begin{figure}[!t]
    \centering
    \includegraphics[width=\textwidth]{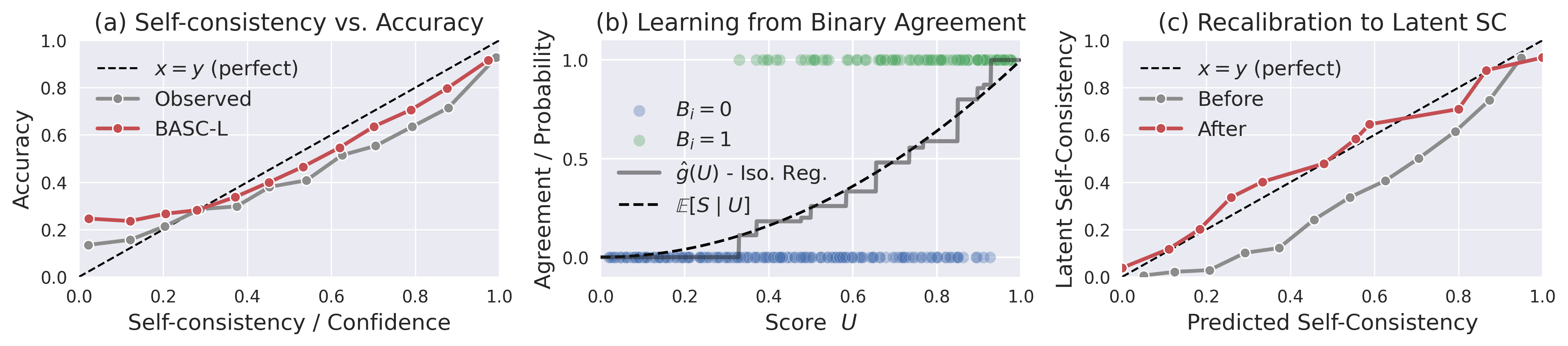}
    \caption{
    Motivation and illustration of BASC.
    \textbf{(a)} Empirical self-consistency and BASC-predicted confidence both track answer correctness across the 36 model--dataset combinations evaluated in our experiments.
    \textbf{(b)} Individual agreement observations \(B\) are binary and noisy, but for any anchor-derived scalar score \(U\),
    \(\mathbb{E}[B\mid U]=\mathbb{E}[S\mid U]\).
    Isotonic regression therefore learns a monotone approximation to
    \(\mathbb{E}[S\mid U]\) from binary agreement observations.
    \textbf{(c)} Binary isotonic regression recalibrates the scalar score with respect to latent self-consistency.
    }
    \label{fig:basc_illustrator}
\end{figure}

\paragraph{\emph{Stage 2: Binary isotonic recalibration.}}
We freeze \(u\) and recalibrate it using independent binary agreement observations.
For each \(x_i\in\mathcal{D}_{\mathrm{iso}}\), draw one anchor
\(Y_i^{(0)}\sim p_\theta(\cdot\mid x_i)\)
and one independent probe
\(Y_i^{(1)}\sim\tilde p_\theta(\cdot\mid x_i)\).
Compute \(U_i=u(x_i,Y_i^{(0)})\) and
\(B_i=\mathbf{1}[A_i^{(1)}\equiv A_i^{(0)}]\).
We fit a one-dimensional isotonic regressor \(\hat g\) to
\(\{(U_i,B_i):x_i\in\mathcal{D}_{\mathrm{iso}}\}\),
assigning each unseen score the fitted value of its nearest observed calibration score
\citep{rossellini2026analysis}
(see Appendix~\ref{sec:app_basc_theory} for implementation details).
By Equation~\ref{eq:score_agreement_identity}, this fits a monotone approximation to
\(\mathbb{E}[S\mid U]\) at the population level.
The final confidence estimator is
\begin{equation}
    \hat c(x,y)=\hat g\!\left(u(x,y)\right).
    \label{eq:final_confidence}
\end{equation}
At deployment, only a single response \(y\) is generated; no probe is needed.

Figure \ref{fig:basc_illustrator} panels (b) and (c) illustrate how BASC uses binary agreement for recalibration. 
In panel (b), each observation records only whether a probe agrees with its anchor, so its value is either zero or one; neither outcome directly reveals the anchor’s self-consistency. 
Across queries, however, the conditional mean of these observations equals the conditional mean of latent self-consistency. 
Isotonic regression pools neighboring score values to fit the step function shown in gray, learning a monotone approximation to this relationship without requiring an accurate self-consistency estimate for any individual query. 
Panel (c) illustrates the resulting change in calibration to self-consistency: after recalibration, reported confidence more closely matches the average latent self-consistency of examples receiving that confidence.

\paragraph{Distribution-free calibration to latent self-consistency.}
Although latent self-consistency \(S\) is never observed directly, BASC inherits a calibration guarantee from binary isotonic regression.
Let \(n_{\mathrm{iso}}=|\mathcal{D}_{\mathrm{iso}}|\), and assume that the Stage~2 and test examples are i.i.d.\ and independent of any data used to construct \(u\).
For a fresh test example, write \(U=u(X,Y^{(0)})\) and \(C=\hat g(U)\).
For readability, for the rest of this section all expectations are over fresh test data with the fitted functions \(u\) and \(\hat g\) held fixed; Appendix~\ref{sec:app_basc_transfer} gives the explicit conditional form.
The binary-isotonic result of \citet{rossellini2026analysis} then gives
\begin{equation}
\mathbb{E}\!\left[
    \left|\mathbb{E}[B\mid C]-C\right|
\right]
=
O_p\!\left(
    n_{\mathrm{iso}}^{-1/6}\sqrt{\log n_{\mathrm{iso}}}
\right),
\label{eq:binary_calibration_bound}
\end{equation}
where \(O_p\) is over the Stage~2 calibration sample.
The left-hand side is the population calibration error of \(C\) with respect to binary agreement \(B\).
The tower property transfers this guarantee exactly to latent self-consistency:
because \(C\) depends only on the anchor and fitted functions,
\[
\mathbb{E}[B\mid C]=\mathbb{E}[S\mid C].
\]
Thus, the population calibration error with respect to \(S\) is identical to that with respect to \(B\), yielding
\begin{equation}
\mathbb{E}\!\left[
    \left|\mathbb{E}[S\mid C]-C\right|
\right]
=
O_p\!\left(
    n_{\mathrm{iso}}^{-1/6}\sqrt{\log n_{\mathrm{iso}}}
\right).
\label{eq:sc_calibration_bound}
\end{equation}
The guarantee depends only on Stage~2 and applies to any frozen scalar score, regardless of its predictive quality, without requiring
\(\mathbb{E}[S\mid U]\) to be monotone.
Appendix~\ref{sec:app_basc_transfer} provides the full argument.
To our knowledge, this is the first distribution-free calibration guarantee for single-generation prediction of LLM self-consistency.

\paragraph{Relationship to correctness calibration.}
The self-consistency guarantee also clarifies when BASC yields well-calibrated confidence with respect to correctness.
Under the same convention of holding the fitted functions fixed, for correctness \(Z\) the triangle inequality gives
\begin{equation}
\left|\mathbb{E}[Z\mid C]-C\right|
\le
\left|\mathbb{E}[S\mid C]-C\right|
+
\left|\mathbb{E}[Z-S\mid C]\right|.
\label{eq:correctness_calibration_gap}
\end{equation}
The left-hand side is the absolute correctness calibration gap at reported confidence \(C\).
Averaged over test examples, Equation~\ref{eq:sc_calibration_bound} controls the first term, i.e., the calibration error with respect to self-consistency, while the second measures how closely self-consistency tracks correctness conditional on reported confidence.
Thus, when this second term is small, BASC can achieve small correctness calibration error using only unlabeled agreement observations.

\section{Experiments}\label{sec:experiments}

We evaluate our proposed approach along four dimensions. \textbf{First,} we test whether self-consistency is in fact a useful unlabeled proxy for answer correctness in reasoning LLMs. 
\textbf{Second,} we ask whether 
binary agreement observations suffice to calibrate a single-generation score.
\textbf{Third,} we examine whether BASC remains reliable under distribution shift between unsupervised training and deployment. 
\textbf{Finally,} we assess whether the confidence estimates are useful for downstream decision-making via selective prediction.
Below we detail our experimental setup; see Appendix \ref{app:exp_details} for additional details.
All parameters and prompts used to produce our data and results are contained in this section and Appendix \ref{app:exp_details}.
Our study relies solely on open source models and datasets, and our code is available at \url{https://github.com/thomaspzollo/basc}.

\paragraph{Reasoning Language Models}
Our models include:
\begin{itemize}
    \item 5 Qwen3 models \citep{yang2025qwen3technicalreport}: 0.6B, 1.7B, 4B-Thinking-2507, 8B, 14B;
    \item 2 Nemotron models \citep{moshkov2025aimo2winningsolutionbuilding, wang2026nemotroncascadescalingcascadedreinforcement}: OpenReasoning-Nemotron-7B, Nemotron-Cascade-8B-Thinking;
    \item 2 DeepSeek models \citep{Guo_2025}: DeepSeek-R1-Distill-Llama-8B, DeepSeek-R1-Distill-Qwen-7B.
\end{itemize}

\paragraph{Datasets}
Our tasks are split between math problem solving and question answering (QA) tasks, following previous work in LLM calibration (e.g., \citet{leng2024taming, luo2025pretrainedllmsecretlyunsupervised,damani2025beyond}).
For math, we use \textbf{GSM8K} \citep{cobbe2021trainingverifierssolvemath} and \textbf{PolyMath} \citep{wang2025polymath}.  PolyMath is a multilingual dataset, and we use data from 8 languages: EN, ES, FR, PT, IT, RU, AR, ID. 
We perform science QA with \textbf{SciQ} \citep{welbl2017crowdsourcingmultiplechoicescience}, and open-domain QA with \textbf{TriviaQA} \citep{joshi2017triviaqalargescaledistantly} and \textbf{WebQuestions (WebQ)} \citep{berant-etal-2013-semantic}.
All models are evaluated on the math datasets.  For free-form QA, we include only the Qwen3 models and the Nemotron-Cascade-8B-Thinking model, as the other 3 models failed to reliably follow the formatting required for automatic extraction for these tasks.
In total, we run 9 models on 2 math datasets and 6 models on 3 QA datasets for a total of 36 combinations (see Table \ref{tab:all_acc} for model/dataset accuracy).
We use 1000 examples per dataset, and average results over 10 random 40/60 calibration/test splits.

\paragraph{Our Method}
We evaluate BASC recalibration (Stage 2) applied to different Stage 1 scores.
\mbox{\textbf{BASC-L}} includes score learning, where we split the calibration set 50/50, and fit an $\ell_2$-regularized binomial-logistic regressor from the model's own last-token hidden state to the $k-1$ agreement outcomes per query, with $k \in \{2, 5\}$.
Using the model's own embeddings integrates seamlessly with existing generation pipelines.
A black-box variant (\textbf{BASC-L-bb}) embeds the prompt and response with an external model, Gemma-3-4B-it; we use $k=5$ for this variant.
\textbf{BASC-TP} uses the pre-existing token probabilities, so all calibration data goes to Stage~2.
Stage 2 recalibration always uses 2 decodes per query.

For each example we sample one anchor response at temperature 0.6 (TopP=0.95, TopK=20), and a pool of 100 probe responses at temperature 0.7 (100 probes allow us to run varied random trials, as well as calculate an accurate empirical self-consistency for each example). Stage~1 subsamples $k-1$ probes per anchor and Stage~2 uses one.
Agreement uses exact matching for math and, for QA, the bidirectional-entailment rule of \citet{kuhn2023semantic} with a DeBERTa-v3 NLI cross-encoder.
Sample code for our method can be seen in Code Box \ref{fig:our_code}.
Given our unsupervised setting, 
we perform no model- or dataset-specific tuning; all method and sampling hyperparameters are fixed across the 36 model/dataset combinations.
We ablate probe sampling temperature in $[0.5, 0.7, 0.9]$ and QA answer matching (compared to string matching) in Appendix \ref{app:ablations}; we find neither choice to make a significant difference to our results.

\paragraph{Evaluation}
We measure expected calibration error using  ECE$_1$ with 12 equal-mass bins (ECE$_2$ is included in Appendix) \citep{guo2017calibration}.
We also measure Brier score (BS) \citep{brier1950verification}, a proper scoring rule for calibration that encourages more discriminative predictions, and AUROC.
While AUROC does not measure calibration, it captures another important quality for useful confidence estimates, that they effectively separate correct and incorrect answers \citep{kadavath2022language}.

\paragraph{Baselines}

The most common baseline for producing LLM confidence estimates is the length-normalized token probability for the full model response (e.g., the geometric mean of the top token probabilities) \citep{kadavath2022language, luo2025pretrainedllmsecretlyunsupervised}.
For reasoning models, to avoid scores being distorted by long thinking chains with many unimportant tokens, one might use the length-normalized token probability of only the tokens associated with the final answer \citep{damani2025beyond}.  Given that both of these baselines are viable without labels or multiple samples, we adopt them (\textbf{Token Probs. (TP)} and \textbf{Answer Probs. (ATP)}) for comparison in our setting.  

Another popular approach to LLM calibration involves having the model directly verbalize its own confidence in its output (\textbf{Verbal Conf.} (VC)) \citep{tian2023just}.  While this does require another full response (including reasoning), which may be prohibitive in resource-constrained settings, we include it because of its prevalence in the LLM calibration literature.
Further, to ensure a strong verbalized confidence baseline, we search for our  prompt among 5 prompts written by GPT-5.2 and 5 written by Gemini-3, and select the prompt with the best results across a selection of models and datasets (see Appendix \ref{sec:verb_conf_imp} for more details and the final prompt).

Finally, we also include results for stronger baselines that violate the resource constraints of no labels and a single test-time generation.  Namely, we include (1) a \textbf{Supervised} baseline where we perform our 2 stage algorithm with ground truth correctness labels and (2) a multiple-sampling baseline measuring empirical \textbf{Test-Time Self-Consistency (TT-SC)} with 100 samples.

\subsection{Main Results}\label{sec:main_results}

Our first experiment addresses two questions: (1) does self-consistency provide a useful unlabeled signal of correctness in reasoning LLMs, and (2) can BASC learn this signal from binary agreement observations alone?
We focus on the standard unsupervised setting, in which the unlabeled calibration data come from the same distribution as the test data \citep{ganin2016domainadversarialtrainingneuralnetworks, luo2025pretrainedllmsecretlyunsupervised}.
Table~\ref{tab:overall_results} reports results averaged over all 36 combinations and ECE$_1$ by model and task family, and Appendix Figures~\ref{fig:res_per_model_qwen}, \ref{fig:res_per_model_non_qwen}, and \ref{fig:res_per_dataset} show results by model and dataset (including standard error measurements).

\begin{table}[!t]
    \centering
    \begin{tabular}{lccccccc}
    \toprule
     & \multicolumn{3}{c}{Average Across 36 Combinations} & \multicolumn{4}{c}{ECE$_1$ (↓) by Model and Task Family} \\
     & ECE$_1$ (↓) & Brier (↓) & AUROC (↑) & Qwen & non-Qwen & Math & QA \\
    \midrule
    TP & 0.197 & 0.222 & 0.655 & 0.198 & 0.195 & 0.102 & 0.293 \\
    ATP & 0.296 & 0.297 & 0.622 & 0.314 & 0.254 & 0.147 & 0.445 \\
    VC & 0.268 & 0.281 & 0.635 & 0.285 & 0.230 & 0.135 & 0.401 \\
    BASC-L (k=2) & \textbf{0.083} & \textbf{0.162} & 0.685 & \textbf{0.089} & \textbf{0.069} & \textbf{0.046} & \textbf{0.119} \\
    BASC-L (k=5) & \textbf{0.084} & \textbf{0.159} & \textbf{0.708} & \textbf{0.091} & \textbf{0.068} & \textbf{0.045} & \textbf{0.123} \\
    BASC-TP & \textbf{0.069} & \textbf{0.167} & 0.647 & \textbf{0.072} & \textbf{0.063} & \textbf{0.045} & \textbf{0.094} \\
    \midrule
    BASC-L-bb (k=5) & \textbf{0.083} & \textbf{0.159} & \textbf{0.705} & \textbf{0.090} & \textbf{0.068} & \textbf{0.045} & \textbf{0.121} \\
    \midrule
    TT-SC & 0.090 & 0.125 & 0.836 & 0.099 & 0.071 & 0.037 & 0.144 \\
    Supervised & 0.051 & 0.147 & 0.732 & 0.053 & 0.046 & 0.038 & 0.064 \\
    \bottomrule
    \end{tabular}
    \caption{
    Calibration results averaged over all 36 model/dataset combinations (left) and ECE$_1$ by model family and task type (right).
    Rows include the unsupervised baselines (token probs. (TP), answer token probs. (ATP), and verbalized confidence (VC)), our methods (BASC-L with $k$ samples per training query, and BASC-TP), a black-box variant with embeddings from an external model (BASC-L-bb), and two strong baselines that break our constraints of no labels and one generation: test-time self-consistency with 100 samples (TT-SC) and supervised recalibration.
    \textbf{Bold results} are significantly better than each of TP, ATP, and VC (paired two-sided Wilcoxon test over the combinations in each column, $p<0.05$).
    BASC-L improves ECE and Brier over all three unsupervised baselines (and AUROC with $k=5$), with ECE$_1$ gains in every model and task family, including in the black-box setting; BASC-TP improves calibration but adds no discrimination to underlying token probabilities.
    }
    \label{tab:overall_results}
\end{table}

With respect to question (1), in our results TT-SC provides a strong unlabeled confidence signal, echoing previous findings \citep{wang2023selfconsistencyimproveschainthought, lyu2025calibrating}.
It matches the ECE of the supervised baseline on math and achieves a better overall Brier score and AUROC, indicating that agreement across sampled responses carries substantial information about correctness.
This supports the central premise of our method: self-consistency is a high-quality target for learning answer-level confidence in reasoning LLMs.
Further quantitative and qualitative measurements of the relationship between self-consistency and correctness for each model and dataset can be found in Appendix \ref{app:exp_results}.

For question (2), BASC-L significantly improves ECE and Brier score over token probabilities, answer probabilities, and verbalized confidence, with ECE$_1$ gains in every model and task family, and with $k=5$ also significantly improves AUROC.
With only two samples per training query, BASC-L ($k=2$) approaches the ECE of test-time self-consistency on math and improves on it for QA, while requiring only a single generation at test time (although it clearly trails on Brier score and AUROC).
BASC-TP, which applies Stage~2 directly to token probabilities, achieves even lower ECE than BASC-L.
However, it inherits the weak discrimination of token probabilities, with lower AUROC and higher Brier score than BASC-L.
This highlights that Stage~2 can substantially improve calibration, while the learned Stage~1 makes the score more informative.
Finally, BASC only requires sampled outputs and can therefore operate without access to model internals.
To test this setting, BASC-L-bb uses embeddings of the prompt and response from an external model, Gemma-3-4B-it, and otherwise runs the same method.
As Table~\ref{tab:overall_results} shows, this black-box variant performs nearly identically to BASC-L on both task types, so our method extends naturally to API-only and proprietary models.

\begin{figure}[!t]
	\centering
	\includegraphics[width=0.9\textwidth]{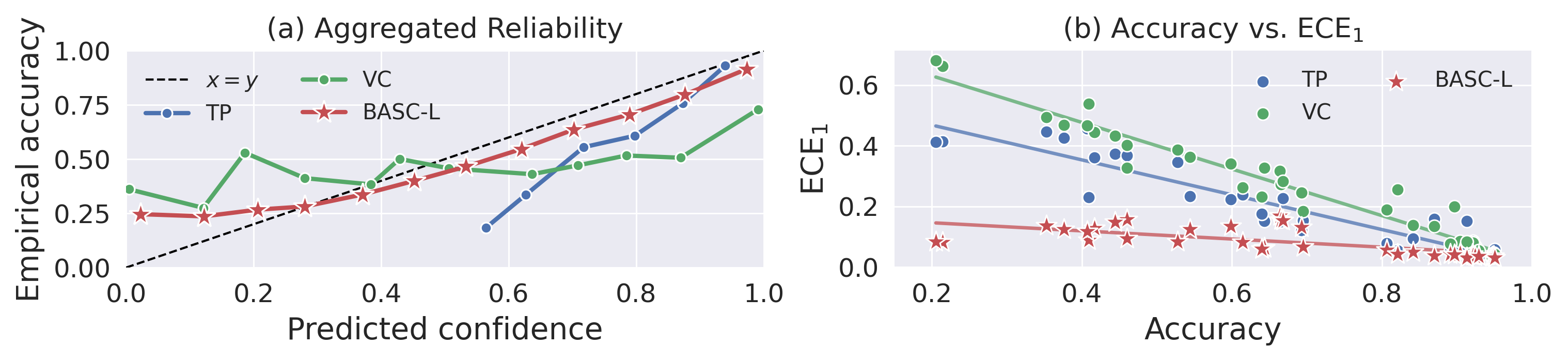}
	\caption{
    \textbf{(a)} Aggregated reliability diagram; perfect calibration lies on the $x=y$ line.
    \textbf{(b)} ECE$_1$ vs.\ accuracy across the 36 model/dataset combinations for our method and baselines.
    }
	\label{fig:main_analysis_1}
\end{figure}

\paragraph{Reliability Diagram} The reliability diagram in Figure~\ref{fig:main_analysis_1}(a), aggregated over all 36 combinations, offers a similar view.
Reliability diagrams visualize calibration by plotting confidence vs.~accuracy for each bin of data; a perfectly calibrated model produces a calibration curve that lies on the line $x=y$, where confidence matches accuracy.
BASC closely tracks the diagonal, while token probabilities are consistently overconfident and verbalized confidence is only loosely related to correctness.

\paragraph{Calibration Error and Accuracy}

Previous work \citep{xiong2024llms} has shown that existing methods (e.g., verbalized confidence) can appear calibrated under ECE because they are usually highly confident and the underlying models are often highly accurate on the tasks under consideration; such calibration breaks down on harder tasks where the model is less accurate.
To test for this effect, Figure~\ref{fig:main_analysis_1}(b) plots accuracy against ECE$_1$ across model/dataset combinations for our method and two unsupervised baselines.
The baselines' calibration error increases substantially as accuracy decreases, whereas our method remains comparatively stable across accuracy levels.
This suggests that its low ECE$_1$ is not merely due to matching average task accuracy with high confidence.

\begin{figure}[!t]
	\centering
	\includegraphics[width=\textwidth]{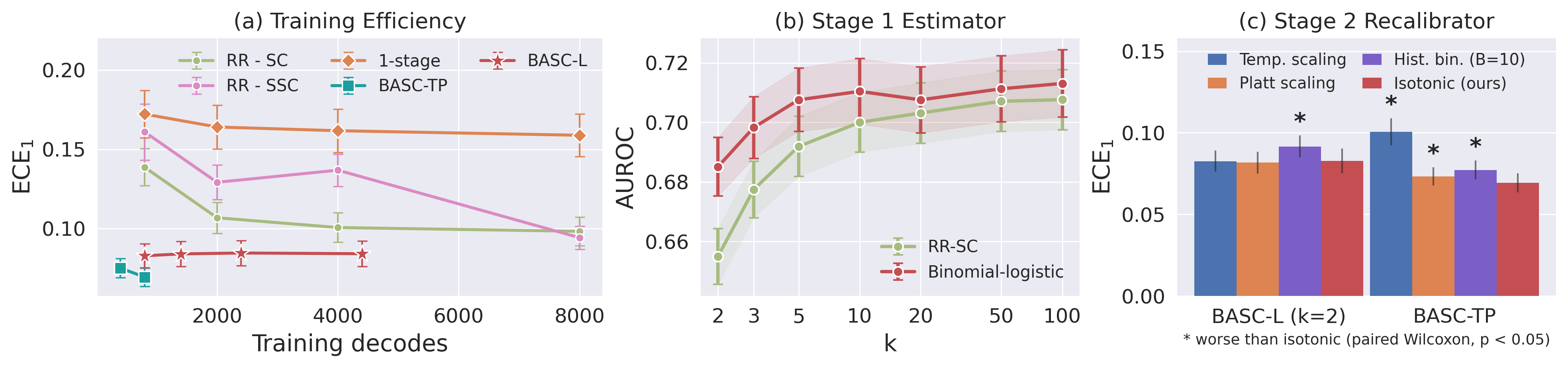}
	\caption{
    \textbf{(a)} ECE$_1$ vs.\ number of full-response training decodes for BASC-L, two single-stage regressors on continuous self-consistency targets (RR-SC, RR-SSC), and BASC-L's Stage-1 model without Stage-2 recalibration (1-stage), at $k \in \{2, 5, 10, 20\}$, and for BASC-TP fit on 200 or 400 queries (2 decodes each). 
    \textbf{(b)} AUROC vs.\ $k$ for BASC's binomial-logistic Stage~1 and ridge regression on empirical self-consistency (RR-SC), each followed by BASC's Stage-2 isotonic recalibration.
    \textbf{(c)} ECE$_1$ of BASC-L ($k=2$) and BASC-TP with isotonic Stage 2 (ours) vs.\ Platt scaling, temperature scaling, and histogram binning (with 10 bins).
    Both stages of BASC matter: with only 800 training decodes, BASC-L has lower ECE$_1$ than the continuous-target baselines with 10$\times$ as many decodes, and about half the ECE$_1$ of its Stage-1 model alone, while the binomial-logistic Stage~1 is more discriminative than the ridge-based RR-SC estimator at small \(k\).
    As the Stage~2 recalibrator, isotonic regression is competitive with alternatives for BASC-L, and significantly outperforms for BASC-TP.
    }
	\label{fig:main_analysis_2}
\end{figure}

\paragraph{Design Decisions}
Figure~\ref{fig:main_analysis_2}(a) plots average ECE$_1$ against the total number of full-response decodes used for calibration.
We compare BASC-L and BASC-TP to two single-stage ridge regressors on BASC-L's embedding features, fit with squared loss to continuous targets: the empirical self-consistency rate (\textbf{RR-SC}) and the confidence-weighted self-consistency of \citet{huang2025efficient} (\textbf{RR-SSC}).
For $k \in \{2, 5, 10, 20\}$, BASC-L uses $200k + 400$ decodes ($k$ per Stage-1 query, 2 per Stage-2 query), while the regressors use $400k$; BASC-TP, which requires no Stage-1 training, uses 2 decodes on each of 200 or 400 queries.
With just 800 decodes, BASC-L is better calibrated than either ridge baseline at 10$\times$ the budget, while BASC-TP performs even better with fewer decodes, demonstrating that BASC can exploit self-consistency without costly per-query estimates.

Panels~\ref{fig:main_analysis_2}(a)--(c) further examine the design choices underlying BASC.
First, a separate held-out recalibration stage is essential: the same binomial-logistic score model fit on all 400 calibration queries without Stage 2 (\textbf{1-stage}, Figure~\ref{fig:main_analysis_2}(a)) has roughly twice BASC-L’s average ECE$_1$ over the 36 model/dataset combinations at every evaluated \(k\).
RR-SC and RR-SSC likewise leave this calibration error unaddressed by fitting self-consistency in a single stage, even if the per-query self-consistency estimation may be more precise.
Second, the binomial-logistic Stage~1 is more discriminative than RR-SC. With Stage~2 held fixed, it achieves significantly higher AUROC when few samples per query are available ($k \le 10$; Figure~\ref{fig:main_analysis_2}(b)).
Thus, BASC’s advantage is not just post-hoc recalibration: its binomial-logistic Stage~1 also learns an informative ranking under limited sampling.
Third, isotonic regression, which carries our distribution-free guarantee, is also empirically strong.
As the Stage~2 recalibrator, isotonic regression performs similarly to the scaling variants and better than histogram binning for BASC-L, and significantly outperforms all three for BASC-TP (Figure~\ref{fig:main_analysis_2}(c)).
Together, these results show that BASC’s design cleanly separates and addresses the two key challenges of confidence estimation, learning an informative score and calibrating it reliably, both without correctness labels.

\paragraph{Ablations} 
In Appendix \ref{app:ablations} we study the effects of two implementation choices: our probe sampling temperature of 0.7, and the use of semantic agreement instead of string matching for the QA tasks (we also discuss the computational costs of semantic matching).  We find that these choices have no significant effects on our results (though a slightly higher sampling temperature seems to help, we do not report that as our main result given the unsupervised setting).

\subsection{Distribution Shift}

Next, we study whether our method is reliable when faced with shifts in the distribution between calibration and test data (noting that this shift nullifies the distribution-free guarantee).
To do so, we evaluate two types of distribution shifts.
First, using the PolyMath dataset, we study distribution shifts across groupings of the 8 languages in our data sample.  
We create 4 different splits based on various language groupings (for example, Romance vs.~non-Romance), and use each split as both training and test data, for a total of 8 experiments (see Appendix \ref{app:ood_exp} for more details on the language splits).
Second, we study distribution shifts across task domains.  For math, we study transfer between PolyMath and GSM8K (2 combinations), and for QA we study transfer among SciQ, TriviaQA, and WebQ (6 combinations).
Results for all models and shifts are summarized in Table \ref{tab:ood_results}, and presented in detail in Appendix \ref{app:ood_exp}; average results with confidence intervals appear in Figure \ref{fig:ood_avg_all_shifts}.

In Table \ref{tab:ood_results}, BASC-L achieves the lowest average ECE$_1$ and Brier score among the single-generation methods in each of the three shift families.
Under language shift, it reduces ECE$_1$/Brier to 0.08/0.14, from 0.14/0.17 for token probabilities; it again performs best under math transfer, and yields the clearest advantage in the challenging QA setting.
It also outperforms ridge regression on empirical self-consistency (RR-SC ($k=10$)) on average within each shift family while using 5$\times$ fewer training decodes.
Despite strong in-distribution ECE, BASC-TP has higher average ECE$_1$ and Brier score than BASC-L within each shift family
and is even worse than raw token probabilities under math transfer, suggesting that the relationship between token probability and agreement changes across datasets.
According to these results, a learned Stage~1 makes BASC meaningfully more robust to shift.

\begin{table}[!t]
    \centering
    \begin{tabular}{lcccccccc}
    \toprule
     & \multicolumn{2}{c}{Language} & \multicolumn{2}{c}{Math} & \multicolumn{2}{c}{QA} & \multicolumn{2}{c}{Average} \\
     & ECE$_1$ & Brier & ECE$_1$ & Brier & ECE$_1$ & Brier & ECE$_1$ & Brier \\
    \midrule
    TP & 0.135 & 0.169 & 0.101 & 0.127 & 0.290 & 0.315 & 0.175 & 0.204 \\
    VC & 0.175 & 0.195 & 0.135 & 0.150 & 0.398 & 0.409 & 0.236 & 0.251 \\
    RR-SC (k=10) & 0.113 & 0.150 & 0.125 & 0.144 & 0.194 & 0.264 & 0.144 & 0.186 \\
    BASC-TP & 0.081 & 0.153 & 0.140 & 0.153 & 0.180 & 0.257 & 0.134 & 0.188 \\
    BASC-L (k=2) & \textbf{0.076} & \textbf{0.139} & \textbf{0.095} & \textbf{0.126} & \textbf{0.157} & \textbf{0.244} & \textbf{0.110} & \textbf{0.170} \\
    \midrule
    TT-SC & 0.045 & 0.079 & 0.035 & 0.066 & 0.141 & 0.184 & 0.074 & 0.109 \\
    \bottomrule
    \end{tabular}
    \caption{
    Calibration results under distribution shift in language (using PolyMath) as well as across domains in math and QA; scores are averaged across models and shifts, lower is better for all metrics.
    Our method offers reliable performance even when calibration data does not match test data.
    }
    \label{tab:ood_results}
\end{table}
\subsection{Selective Prediction}\label{subsec:sel_pred}

Selective prediction uses confidence estimates to decide which examples to answer and which to abstain on, with the goal of improving accuracy on the selected examples by abstaining on the most uncertain cases \citep{geifman2017selective, geifman2019selectivenet, yadkori2024mitigating}. 
Although this is not strictly dependent on calibration (e.g., selective accuracy is invariant to strictly increasing transformations of the confidence scores), the ability to order examples from least to most likely to be correct is another desirable quality in a confidence estimator.

To study whether our method is useful in this way, we use the same experimental setup as Section \ref{sec:main_results} (40/60 data split), using the unlabeled calibration data to train our confidence estimation model.
Using our method and baselines, we rank test examples by confidence and evaluate abstention rates from 5\% to 90\%, measuring the accuracy on the non-abstained (selected) set.
In this experiment we focus on BASC-L, as BASC-TP does not learn a new ranking.
Selective accuracy is averaged over 360 trials (36 combinations and 10 seeds), and we also analyze the rejected and selected sets by comparing their average confidence and realized accuracy; this lets us test not only whether a method selects good examples, but also whether it provides a calibrated picture of the quality of both the rejected and accepted examples. 

\begin{figure}[!t]
	\centering
	\includegraphics[width=\textwidth]{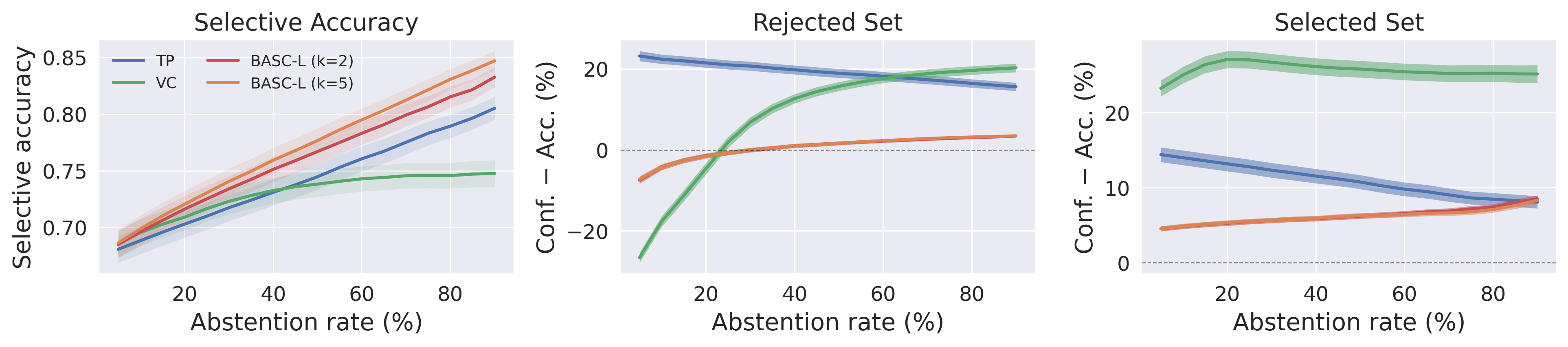}
	\caption{
    On the left, results for selective prediction using our method and baselines.
    The X-axis reflects the percent of examples on which the system abstains, and the y-axis reflects the corresponding accuracy for retained/selected examples.
    In the middle and right plots, we examine the difference between the average confidence and accuracy of the abstained and selected examples for each method at each level of abstention.
    Taken together, these plots highlight that our method enables selection that brings gains in accuracy while also allowing the decision-maker to understand the likely accuracy on both rejected and accepted predictions.
    }
	\label{fig:selection_results}
\end{figure}

Figure~\ref{fig:selection_results} (including standard error intervals) shows that our method yields the strongest overall selective prediction performance among the methods available in our setting. 
In Figure~\ref{fig:selection_results} (left), abstaining according to our confidence estimates produces substantially larger gains in retained-set accuracy across the full range of abstention rates, whereas the baselines improve more slowly, and VC plateaus around 50\%. 
Moreover, the middle and right plots show that, for both the rejected and selected sets, our method's average confidence tracks realized accuracy much more closely than the alternatives, enabling a decision-maker to understand the likely accuracy of both rejected and accepted predictions.
Appendix Figures~\ref{fig:sel_by_dataset} and~\ref{fig:sel_by_model} show that these trends hold across model families and datasets.

\vspace{-6pt}
\section{Conclusion}

We find that calibrated confidence for reasoning LLMs need not require either labeled calibration data (and possibly compute-intensive RL training) or expensive repeated sampling.
Importantly, this approach is not limited to open-weight models: our black-box results suggest that the same strategy can remain effective even when the underlying reasoning model is only accessible through a proprietary interface.
Our results also highlight that verbalized confidence with reasoning LLMs remains a weak signal for calibration, at least without model training or extensive prompt optimization.

\paragraph{Limitations}
BASC is only guaranteed to offer calibration with respect to latent self-consistency, and not actual correctness.
While our experiments show that self-consistency (and BASC-predicted confidence) do track correctness closely across a wide range of models and datasets, our method can only be as good as the self-consistency signal it learns: if agreement among samples is weakly related to correctness, or models are consistently wrong with high internal agreement, the learned predictor may inherit those failures. 
Finally, our distribution-free bound is not tight at the calibration set sizes used in our experiments. Nevertheless, it establishes that population calibration error to latent self-consistency goes to zero as the number of calibration queries grows, even with only two responses per query.


\section*{Acknowledgments}
We thank ONR Grant N00014-23-1-2436 for its generous support.  
This work is supported by the funds provided by the National Science Foundation and by DoD OUSD (R\&E) under Cooperative Agreement PHY-2229929 (The NSF AI Institute for Artificial and Natural Intelligence).

\bibliography{iclr2027_conference}

\clearpage

\appendix

\section{Additional Method Details}
\label{sec:app_basc_theory}

\subsection{Transfer from binary agreement to latent self-consistency.}
\label{sec:app_basc_transfer}

We state the binary-isotonic calibration result used in the main text and show why it transfers exactly from observable agreement \(B\) to latent self-consistency \(S\).

\paragraph{Sampling assumptions and notation.}
Condition on the frozen score function \(u\).
If \(u\) is learned, all data and randomness used to fit or select it are independent of the Stage~2 calibration sample and fresh test example; for a pre-existing score, \(u\) is fixed from the outset.
Stage~2 and test prompts are drawn independently from the same distribution, and their responses are generated using the fixed anchor and probe sampling procedures of Section~\ref{sec:method}, with independent sampling randomness across examples.
Consequently, conditional on \(u\), the Stage~2 observations
\(\{(U_i,B_i)\}_{i=1}^{n_{\mathrm{iso}}}\)
and a fresh test pair \((U,B)\) are i.i.d., with binary \(B\).
The fitted isotonic regressor \(\hat g\) depends only on \(u\) and the Stage~2 observations, and is therefore independent of the fresh test example conditional on \(u\).

For the fresh example, write
\[
    U=u(X,Y^{(0)}),\qquad
    S=s^\star(X,A^{(0)}),\qquad
    C=\hat g(U),
\]
and let \(B=\mathbf{1}[A^{(1)}\equiv A^{(0)}]\) denote agreement with an independent probe.
This test probe is introduced only for the analysis; BASC does not sample it at deployment.
For a target \(T\in\{B,S,Z\}\), define the population calibration error
\[
    \operatorname{CE}_T(\hat g,u)
    :=
    \mathbb{E}\!\left[
        \left|\mathbb{E}[T\mid C,\hat g,u]-C\right|
        \,\middle|\,\hat g,u
    \right].
\]
Both expectations are over fresh test data, with the fitted functions held fixed.


\paragraph{Binary-isotonic calibration.}
For the nearest-neighbor extension described below, Theorem~2 of
\citet{rossellini2026analysis} implies that there exists a universal
constant \(K>0\) such that, for every \(n_{\mathrm{iso}}\ge2\) and
\(\delta\in(0,1)\),
\begin{equation}
    \mathbb{P}\!\left(
        \operatorname{CE}_B(\hat g,u)
        \le
        K n_{\mathrm{iso}}^{-1/6}
        \sqrt{\log(n_{\mathrm{iso}}/\delta)}
        \,\middle|\,u
    \right)
    \ge 1-\delta.
    \label{eq:app_binary_calibration_bound}
\end{equation}
The probability is over the Stage~2 calibration sample, conditional
on the frozen score \(u\); \(K\) does not depend on \(u\),
\(n_{\mathrm{iso}}\), \(\delta\), or the data distribution.
Consequently,
\[
    \operatorname{CE}_B(\hat g,u)
    =
    O_p\!\left(
        n_{\mathrm{iso}}^{-1/6}\sqrt{\log n_{\mathrm{iso}}}
    \right),
\]
where \(O_p\) is taken over the same conditional sampling distribution.
Subject to the sampling assumptions above, the guarantee is
distribution-free: it places no further restrictions on the joint
distribution of \((U,B)\) beyond \(B\in\{0,1\}\).
In particular, it requires neither predictive accuracy of \(u\)
nor monotonicity of \(\mathbb{E}[B\mid U]\).
The bound concerns population calibration on fresh examples rather
than empirical calibration on the fitting sample.


\paragraph{Exact transfer to latent self-consistency.}
Independence of the fresh example from the data used to construct \(u\) and \(\hat g\), together with Equation~\ref{eq:agreement_expectation}, implies
\begin{equation}
    \mathbb{E}\!\left[
        B\mid X,Y^{(0)},\hat g,u
    \right]
    =
    s^\star(X,A^{(0)})
    =
    S.
    \label{eq:app_basc_anchor}
\end{equation}
Because \(C=\hat g(u(X,Y^{(0)}))\) is measurable with respect to
\((X,Y^{(0)},\hat g,u)\), the tower property gives
\begin{align}
    \mathbb{E}[B\mid C,\hat g,u]
    &=
    \mathbb{E}\!\left[
        \mathbb{E}[B\mid X,Y^{(0)},\hat g,u]
        \,\middle|\,C,\hat g,u
    \right] \\
    &=
    \mathbb{E}[S\mid C,\hat g,u].
    \label{eq:app_basc_transfer}
\end{align}
Thus the calibration discrepancies with respect to \(B\) and \(S\) agree almost surely, and
\[
    \operatorname{CE}_S(\hat g,u)
    =
    \operatorname{CE}_B(\hat g,u).
\]
Combining this exact identity with Equation~\ref{eq:app_binary_calibration_bound} proves Equation~\ref{eq:sc_calibration_bound}.
The transfer introduces no additional approximation error and does not require observing or accurately estimating \(S\) for any calibration query.

\paragraph{Relationship to correctness calibration.}
Let \(Z\in\{0,1\}\) denote correctness of the fresh anchor answer.
Linearity of conditional expectation and the triangle inequality give
\[
    \left|\mathbb{E}[Z\mid C,\hat g,u]-C\right|
    \le
    \left|\mathbb{E}[Z-S\mid C,\hat g,u]\right|
    +
    \left|\mathbb{E}[S\mid C,\hat g,u]-C\right|.
\]
Averaging over the fresh test example yields
\begin{align}
    \operatorname{CE}_Z(\hat g,u)
    &\le
    \operatorname{CE}_S(\hat g,u)
    +\Delta_{Z,S}(\hat g,u),
    \label{eq:app_basc_correctness_gap}\\
    \Delta_{Z,S}(\hat g,u)
    &:=
    \mathbb{E}\!\left[
        \left|\mathbb{E}[Z-S\mid C,\hat g,u]\right|
        \,\middle|\,\hat g,u
    \right].
    \nonumber
\end{align}
BASC controls the self-consistency calibration term, whereas
\(\Delta_{Z,S}\) measures the remaining discrepancy between correctness and self-consistency conditional on reported confidence.
For example, if
\(\left|\mathbb{E}[Z-S\mid C,\hat g,u]\right|\le\varepsilon\)
almost surely, then
\(\operatorname{CE}_Z(\hat g,u)\le\operatorname{CE}_S(\hat g,u)+\varepsilon\).
Such a condition is an additional assumption about the quality of self-consistency as a correctness proxy; it is not implied by binary agreement supervision.

\subsection{Isotonic fitting and out-of-sample prediction.}
\label{sec:app_basc_isotonic}

Stage~2 fits \(\hat g\) using the pool-adjacent-violators algorithm (PAVA).
After grouping identical scores, let
\(v_1<\cdots<v_m\) denote the distinct observed score values, with \(n_j\)
observations and mean agreement \(\bar B_j\) at \(v_j\).
PAVA solves
\begin{equation}
    \min_{c_1\leq\cdots\leq c_m}
    \sum_{j=1}^{m}
    n_j\left(\bar B_j-c_j\right)^2.
\end{equation}
Starting from singleton score blocks, PAVA repeatedly merges adjacent blocks
whose fitted means violate the monotonicity constraint.
At termination, each block receives its pooled empirical agreement rate,
and the fitted block values are nondecreasing and lie in \([0,1]\).

For an unseen score \(U\), we use the nearest-neighbor extension considered by
\citet{rossellini2026analysis}: \(U\) receives the fitted value associated with
its nearest observed score, and hence with that score's fitted PAVA block.
Equidistant ties are resolved using a fixed rule; scores outside the observed range receive the corresponding endpoint value.
This yields a piecewise-constant predictor whose outputs are fitted block means.

Although PAVA constrains \(\hat g\) to be monotone, the calibration guarantee
does not assume that the true regression function
\[
    m(U)=\mathbb{E}[B\mid U]=\mathbb{E}[S\mid U]
\]
is monotone.
When \(m\) is non-monotone, isotonic regression provides a monotone
approximation to \(m\).
The guarantee concerns calibration of the reported confidence \(C=\hat g(U)\),
as measured by \(\operatorname{CE}_B(\hat g,u)=\operatorname{CE}_S(\hat g,u)\),
rather than pointwise recovery of \(m(U)\).
Calibration alone therefore does not ensure an informative ranking of examples.

\clearpage

\section{Additional Experiment Details}\label{app:exp_details}

Here we report additional experiment details.  All models and datasets are available for public download from \url{https://huggingface.co/}.  We use the vLLM \citep{kwon2023efficient} library to produce all language model outputs and embeddings. 
For fair comparison, all models and datasets use the exact same prompt, sampling parameters, and other setup.

All models receive the following prompt:
\begin{quote}
\texttt{Please reason step by step, and put your final answer within \textbackslash boxed\{\}.\newline Question: \{question\}}
\end{quote}

For each example we sample one anchor response at temperature 0.6 (TopP=0.95, TopK=20), the recommended inference setting for these models. 
We also sample a pool of 100 probe responses at temperature 0.7; Stage 1 subsamples $k-1$ probes per anchor and Stage 2 uses one.
For training, we calculate binary agreement using exact matching for math and semantic matching \citep{kuhn2023semantic} for QA tasks.

For each dataset, we sample 1000 examples.  For PolyMath, we sample 125 examples from each of 8 languages (EN, ES, FR, PT, IT, RU, AR, ID); we use the ``low'' split in order to facilitate easy evaluation via matching.

\subsection{Our Algorithm}

See Code Box \ref{fig:our_code} for sample code for training our unsupervised confidence calibrator and applying it to new test inputs.

\begin{listing}[!t]
\begin{lstlisting}[language=Python]
import numpy as np
from sklearn.pipeline import Pipeline
from sklearn.preprocessing import StandardScaler
from sklearn.linear_model import LogisticRegression
from sklearn.isotonic import IsotonicRegression

# Three-way split: Stage-1 fit (s1), Stage-2 fit (s2), test.
# X_*: anchor embeddings; A_*: anchor answers; P_*: probe answers.
# k = 2 (one anchor + one probe per query), headline setting.

# 1. Stage-1 supervision on split s1: binomial (m, c) from
# probe-vs-anchor agreements.
m_s1 = np.full(len(X_s1), k - 1)          # probes per query
c_s1 = agree(A_s1, P_s1).sum(axis=1)      # agreements per query

# Expand each query into (pos, neg) pseudo-rows weighted by
# (c, m - c); LogisticRegression then maximizes the binomial
# log-likelihood.
X_exp = np.concatenate([X_s1, X_s1])
y_exp = np.concatenate([np.ones(len(X_s1)), np.zeros(len(X_s1))])
w_exp = np.concatenate([c_s1, m_s1 - c_s1])
keep = w_exp > 0

stage1 = Pipeline([
    ("scaler", StandardScaler()),
    ("clf", LogisticRegression(C=1.0, max_iter=5000)),
])
stage1.fit(X_exp[keep], y_exp[keep],
           clf__sample_weight=w_exp[keep])

# 2. Stage-2 supervision on split s2: independent (U, B) pairs
# where U = stage1(X_s2) and B is a single binary probe agreement.
U_s2 = stage1.predict_proba(X_s2)[:, 1]
B_s2 = agree(A_s2, P_s2).ravel().astype(int)

# Distribution-free binary isotonic recalibrator. We fit PAVA on
# (U_s2, B_s2) and extend the fitted step function to unseen
# scores by snapping each to the value at its nearest fitted knot.
stage2_iso = IsotonicRegression(
    y_min=0.0, y_max=1.0, out_of_bounds="clip",
)
stage2_iso.fit(U_s2, B_s2)
kx = np.asarray(stage2_iso.X_thresholds_, dtype=float)
ky = np.asarray(stage2_iso.y_thresholds_, dtype=float)

def stage2_predict(scores):
    U = np.asarray(scores, dtype=float)
    j = np.abs(kx - U[:, None]).argmin(axis=1)   # nearest knot
    return np.clip(ky[j], 0.0, 1.0)

# 3. Confidence estimate on test inputs.
U_test    = stage1.predict_proba(X_test)[:, 1]
conf_test = stage2_predict(U_test)
\end{lstlisting}
\caption{
Code for training and applying BASC, our two-stage unsupervised
confidence calibrator. 
}
\label{fig:our_code}
\end{listing}

\subsection{Evaluation}

ECE$_1$ and ECE$_2$ are calculated using the software package released with \citep{kumar2020verifieduncertaintycalibration}.\footnote{\url{https://github.com/p-lambda/verified_calibration}}  Brier score and AUROC are calculated using the sklearn python package \citep{scikit-learn}.

\paragraph{Answer extraction and matching}
The answer of a response is the content of its last \texttt{\textbackslash boxed\{\}} span, trimmed; a response with no such span yields an empty answer, which is scored incorrect.
For math (GSM8K, PolyMath), both correctness (answer vs.\ gold) and anchor--probe agreement are exact string equality.
For QA (SciQ, TriviaQA, WebQ), correctness is a case-insensitive two-way substring match (either string containing the other) against the gold answer or its alias list, after removing any \texttt{\textbackslash text\{\}} wrapper and LaTeX backslashes.
QA agreement between two model answers uses mutual entailment \citep{kuhn2023semantic}: the \texttt{cross-encoder/nli-deberta-v3-base} NLI model is run in both directions on the strings ``Question: \{q\} Answer: \{a\}'', and the answers agree only if both directions are labeled entailment.
Appendix~\ref{app:ablations} ablates this choice against exact-match agreement on QA.

\subsection{Baselines}

Here we provide more details for baseline implementations.

\subsubsection{Token Probabilities}

For the token probabilities baselines, we extract top token probabilities using vLLM's built-in API.  We length-normalize the probabilities by computing their geometric mean, as is typical in the language modeling uncertainty literature \citep{malinin2021uncertainty}.

For the Token Probs. baseline we compute this for the entire response produced by the LLM.  For the Answer Probs. baseline, we do so for the tokens inside \textbackslash{boxed}\{\}.

\subsubsection{Verbalized Confidence}\label{sec:verb_conf_imp}

Empirical research has repeatedly shown that when LLMs are asked to verbalize their confidence in a response, the resulting estimates are often poorly calibrated, and good calibration is often incidental \citep{xiong2024llms, zhou2023navigating}.  
For example, in recent work, \citet{damani2025beyond} found that LLMs trained with reinforcement learning perform poorly at verbalizing confidence, usually worse than just using the token probabilities (see Table 1 in the paper).

However, in an effort to present the strongest baselines possible, we search for our prompt among 5 prompts written by GPT-5.2 and 5 written by Gemini-3 (both via their respective web interface).  We rank all prompts based on our evaluation measures, as well as the rate at which the model fails to produce a well-formatted confidence estimate using the prompt (Imputed, where lower implies better).  When the model fails to produce a well-formatted confidence estimate, we impute the estimate as the average estimate across the test set where estimates are provided (also a strong implementation choice for the baseline, as the test set will generally not all be available at once).  

We run our optimization across 4 models (Qwen3-1.7B/8B, Nemotron-Cascade-8B-Thinking, DeepSeek-R1-Distill-Llama-8B) and 2 datasets (GSM8K, TriviaQA), excluding DeepSeek-R1-Distill-Llama-8B/TriviaQA.
Results for the 10 prompts are shown in Table \ref{tab:vc_prompt_comp}, where GPT-5.2 produces prompts 0-4, and Gemini-3 produces prompts 5-9.
The best-performing prompt, which we use in our experiments, is shown in Prompt Box~\ref{fig:vc_prompt_final}.
All 10 prompts can be found in our code.

\begin{table}[!t]
    \centering
    \begin{tabular}{lcccccc}
    \toprule
    Prompt & ECE$_1$ & ECE$_2$ & BS & Imputed & AUROC & Rank \\
    \midrule
    6 & 0.201 & 0.224 & 0.211 & 0.008 & 0.649 & 1 \\
    8 & 0.195 & 0.221 & 0.214 & 0.009 & 0.628 & 2 \\
    9 & 0.204 & 0.227 & 0.216 & 0.007 & 0.638 & 3 \\
    4 & 0.204 & 0.228 & 0.216 & 0.012 & 0.629 & 4.5 \\
    5 & 0.202 & 0.230 & 0.216 & 0.014 & 0.642 & 4.5 \\
    0 & 0.203 & 0.228 & 0.220 & 0.008 & 0.624 & 6 \\
    3 & 0.202 & 0.226 & 0.218 & 0.016 & 0.630 & 7 \\
    1 & 0.205 & 0.229 & 0.215 & 0.015 & 0.636 & 8 \\
    2 & 0.209 & 0.227 & 0.218 & 0.019 & 0.621 & 9 \\
    7 & 0.227 & 0.249 & 0.233 & 0.013 & 0.608 & 10 \\
    \bottomrule
    \end{tabular}
    \caption{Results for optimization of verbalized confidence baseline.}
    \label{tab:vc_prompt_comp}
\end{table}

\captionsetup[listing]{name=Prompt Box}

\begin{listing}[!t]
\begin{lstlisting}[style=promptbox]
You are a fact-checking and calibration model.
Task: determine the likelihood that the ANSWER accurately solves the given QUESTION.
Output format (STRICT):
1) Use at most FOUR concise sentences to explain your reasoning.
2) On a NEW LINE, output ONLY the probability using the format \boxed{p} (p is a decimal in [0,1]).
   - Format examples: \boxed{0.18}, \boxed{0.42}, \boxed{0.66}, \boxed{0.99}
   - Do not reuse these example scores; instead, be as accurate as possible for this specific case.
3) Do NOT write any text after the \boxed{...} line.
\end{lstlisting}
\caption{
Verbalized confidence prompt used in our experiments.
}
\label{fig:vc_prompt_final}
\end{listing}



\clearpage
\section{Base Models for Weak Supervision}\label{sec:app_base_targets}

Recent work on unsupervised calibration for LLMs relies on the confidence estimates from a paired pre-trained or base model for calibration \citep{luo2025pretrainedllmsecretlyunsupervised, tan2026basecalunsupervisedconfidencecalibration}. 
This approach is motivated by the common empirical finding that models trained with reinforcement learning display worse calibration than the base models from which they are derived \citep{kadavath2022language, leng2024taming}.
However, these methods are not suitable for our setting.
\citet{luo2025pretrainedllmsecretlyunsupervised} is focused on multiple choice tasks, and does not study the problem for open-ended generation.

More importantly, these methods both rely on using the base model to assign confidence estimates to the output of a post-trained model.
This design may fail conceptually with reasoning models, though: long chain-of-thought outputs and greatly enhanced capabilities can create a substantial distribution shift for the paired base model, meaning their token probabilities may no longer be reflective of task correctness. 
To verify this hypothesis, we run a set of experiments with all 5 Qwen3 models (and their paired base model versions) where we follow \citet{luo2025pretrainedllmsecretlyunsupervised} and \citet{tan2026basecalunsupervisedconfidencecalibration} in extracting the base model confidence for the output produced by the post-trained model.
In order to present the strongest case possible, we explore the method both with (\textbf{Base Response Probs.}) and without (\textbf{Base Answer Probs.}) reasoning chains included.

Results by model and dataset are presented in Tables \ref{tab:app_base_targets_model} and \ref{tab:app_base_targets_dataset}.
We compare the results to \textbf{self-consistency} scores with 100 samples.
In our setting, base-model-derived confidence scores perform poorly compared to self-consistency across all Qwen3 models and datasets.  Given the conceptual gap and empirical failure, and our goal to present clear and concise results, we do not include these methods in our main experiments.

\begin{table}[!t]
    \centering
    \begin{tabular}{llrrr}
    \toprule
    Model & Method & ECE$_2$ & Brier & AUROC \\
    \midrule
    Qwen3-0.6B & Base Answer Probs. & 0.410 & 0.392 & 0.471 \\
    Qwen3-0.6B & Base Response Probs. & 0.221 & 0.241 & 0.390 \\
    Qwen3-0.6B & Test-time SC & 0.081 & 0.125 & 0.852 \\
    \midrule
    Qwen3-1.7B & Base Answer Probs. & 0.568 & 0.543 & 0.458 \\
    Qwen3-1.7B & Base Response Probs. & 0.273 & 0.256 & 0.344 \\
    Qwen3-1.7B & Test-time SC & 0.130 & 0.142 & 0.828 \\
    \midrule
    Qwen3-4B-Thinking-2507 & Base Answer Probs. & 0.635 & 0.595 & 0.496 \\
    Qwen3-4B-Thinking-2507 & Base Response Probs. & 0.325 & 0.271 & 0.410 \\
    Qwen3-4B-Thinking-2507 & Test-time SC & 0.140 & 0.137 & 0.819 \\
    \midrule
    Qwen3-8B & Base Answer Probs. & 0.664 & 0.626 & 0.482 \\
    Qwen3-8B & Base Response Probs. & 0.280 & 0.244 & 0.366 \\
    Qwen3-8B & Test-time SC & 0.124 & 0.133 & 0.808 \\
    \midrule
    Qwen3-14B & Base Answer Probs. & 0.638 & 0.574 & 0.507 \\
    Qwen3-14B & Base Response Probs. & 0.291 & 0.250 & 0.434 \\
    Qwen3-14B & Test-time SC & 0.120 & 0.129 & 0.820 \\
    \bottomrule
    \end{tabular}
    \caption{Comparing potential sources of weak labels for unsupervised calibration; results are by model, averaged across 5 datasets.}
    \label{tab:app_base_targets_model}
\end{table}

\begin{table}[!t]
    \centering
    \begin{tabular}{llccc}
    \toprule
    Dataset & Method & ECE$_2$ & Brier & AUROC \\
    \midrule
    GSM8K & Base Answer Probs. & 0.807 & 0.732 & 0.490 \\
    GSM8K & Base Response Probs. & 0.401 & 0.235 & 0.385 \\
    GSM8K & Test-time SC & 0.037 & 0.051 & 0.795 \\
    \midrule
    Polymath & Base Answer Probs. & 0.675 & 0.581 & 0.544 \\
    Polymath & Base Response Probs. & 0.362 & 0.237 & 0.353 \\
    Polymath & Test-time SC & 0.064 & 0.068 & 0.903 \\
    \midrule
    SciQ & Base Answer Probs. & 0.608 & 0.595 & 0.397 \\
    SciQ & Base Response Probs. & 0.245 & 0.287 & 0.418 \\
    SciQ & Test-time SC & 0.168 & 0.185 & 0.817 \\
    \midrule
    Trivia QA & Base Answer Probs. & 0.479 & 0.469 & 0.504 \\
    Trivia QA & Base Response Probs. & 0.245 & 0.263 & 0.357 \\
    Trivia QA & Test-time SC & 0.123 & 0.130 & 0.894 \\
    \midrule
    WebQ & Base Answer Probs. & 0.346 & 0.352 & 0.480 \\
    WebQ & Base Response Probs. & 0.136 & 0.240 & 0.429 \\
    WebQ & Test-time SC & 0.204 & 0.232 & 0.718 \\
    \bottomrule
    \end{tabular}
    \caption{Comparing potential sources of weak labels for unsupervised calibration; results are by dataset, averaged across 5 Qwen3 models.}
    \label{tab:app_base_targets_dataset}
\end{table}

\clearpage
\section{Additional Results}\label{app:exp_results}

Here we include additional results for all experiments.

\subsection{Main Results}

\begin{figure}[!t]
	\centering
	\includegraphics[width=\textwidth]{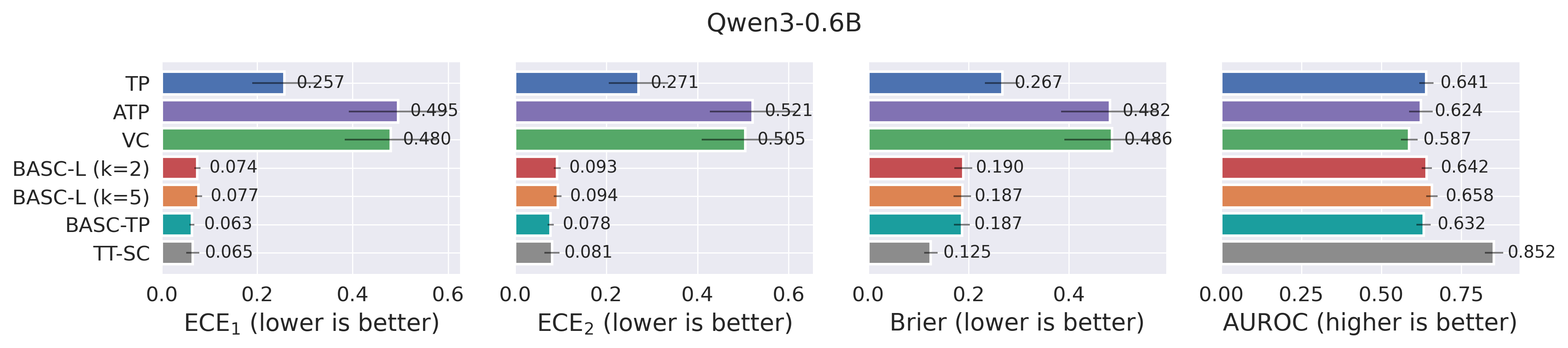}
    \includegraphics[width=\textwidth]{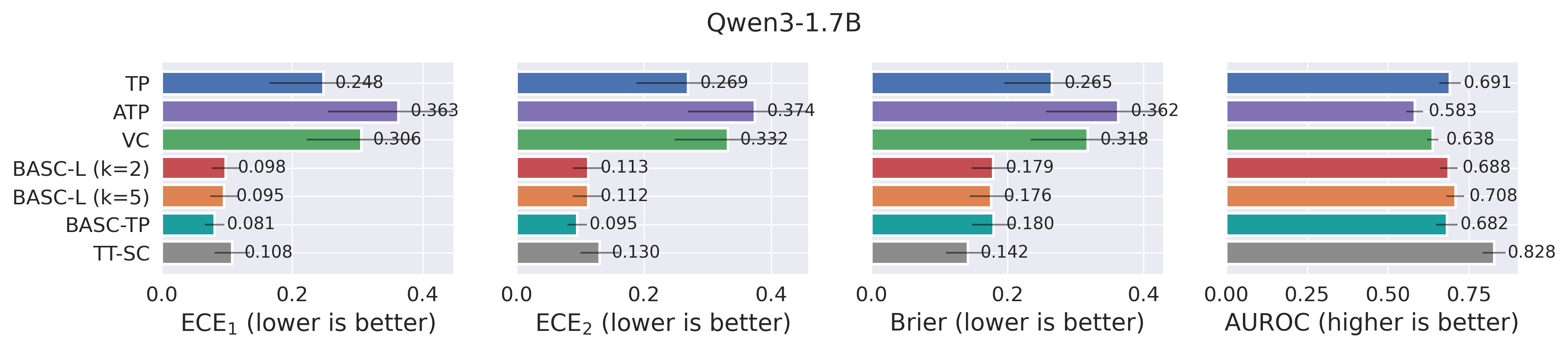}
    \includegraphics[width=\textwidth]{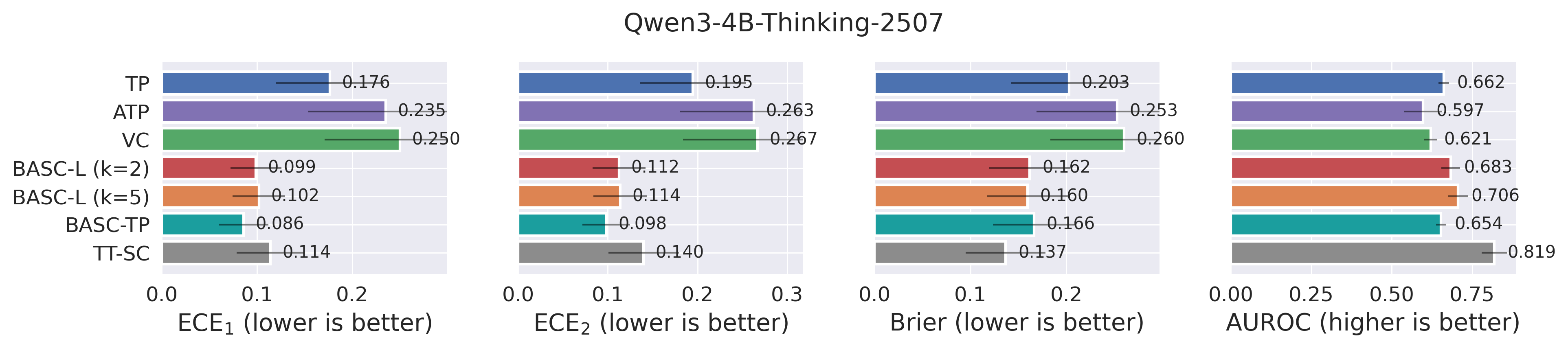}
    \includegraphics[width=\textwidth]{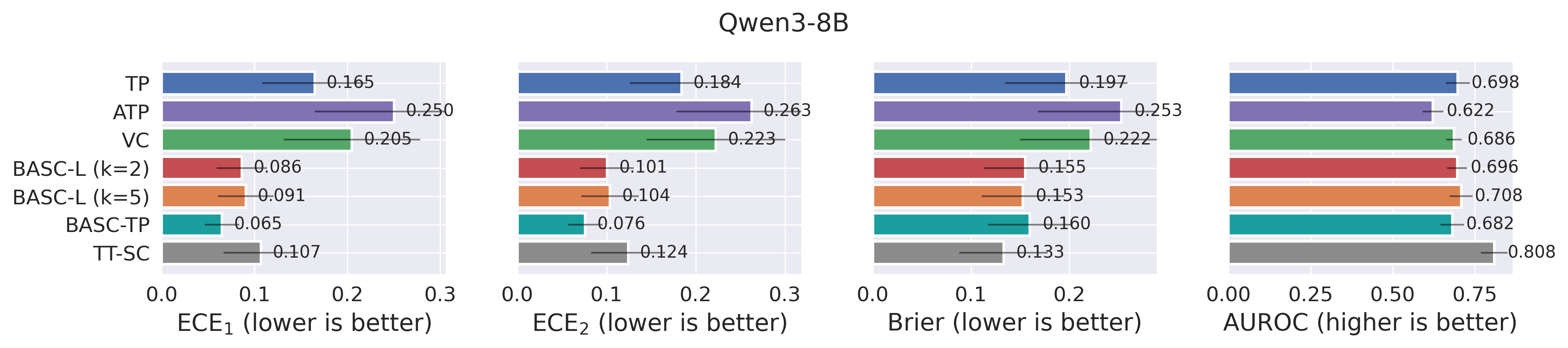}
    \includegraphics[width=\textwidth]{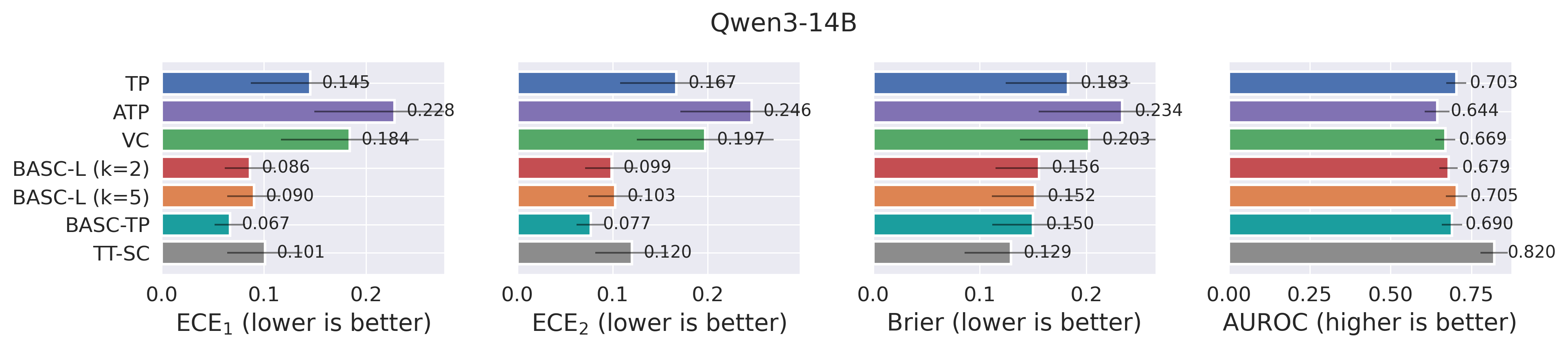}
	\caption{
    Main results for Qwen models.
    }
	\label{fig:res_per_model_qwen}
\end{figure}

\begin{figure}[!t]
	\centering
    \includegraphics[width=\textwidth]{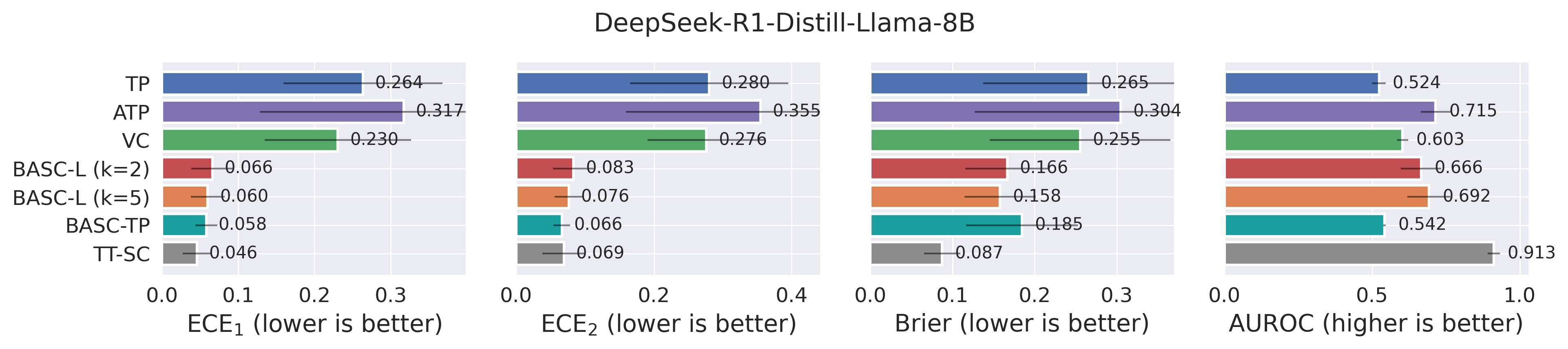}
    \includegraphics[width=\textwidth]{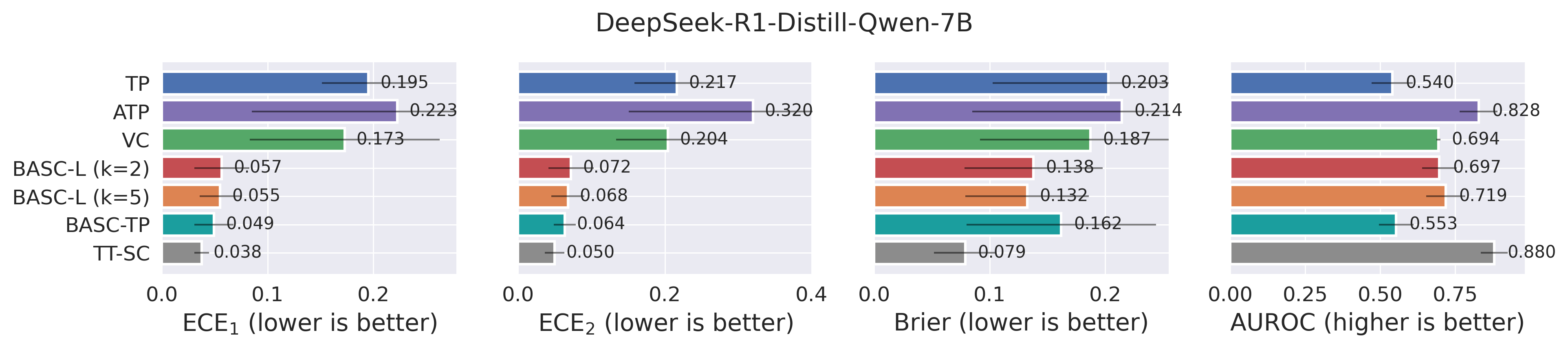}
    \includegraphics[width=\textwidth]{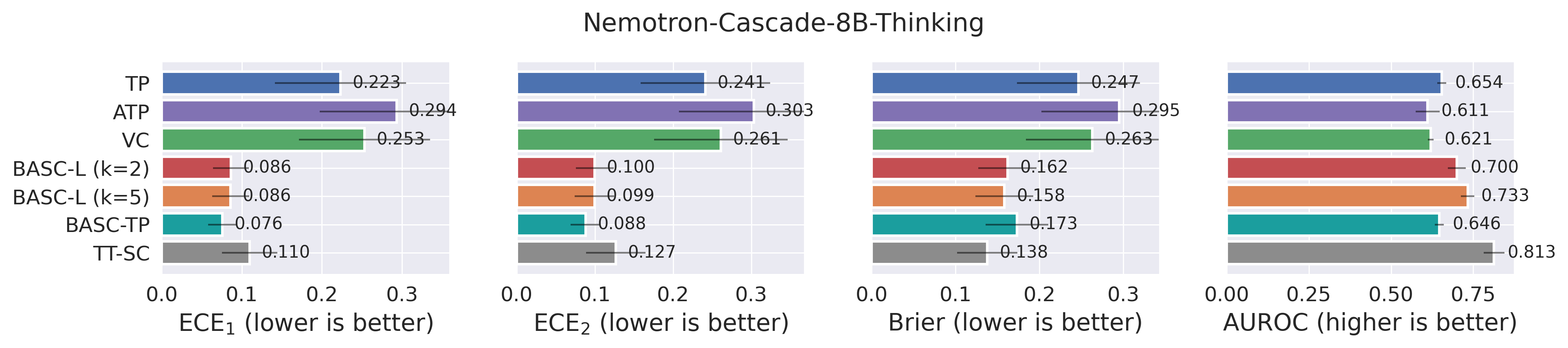}
    \includegraphics[width=\textwidth]{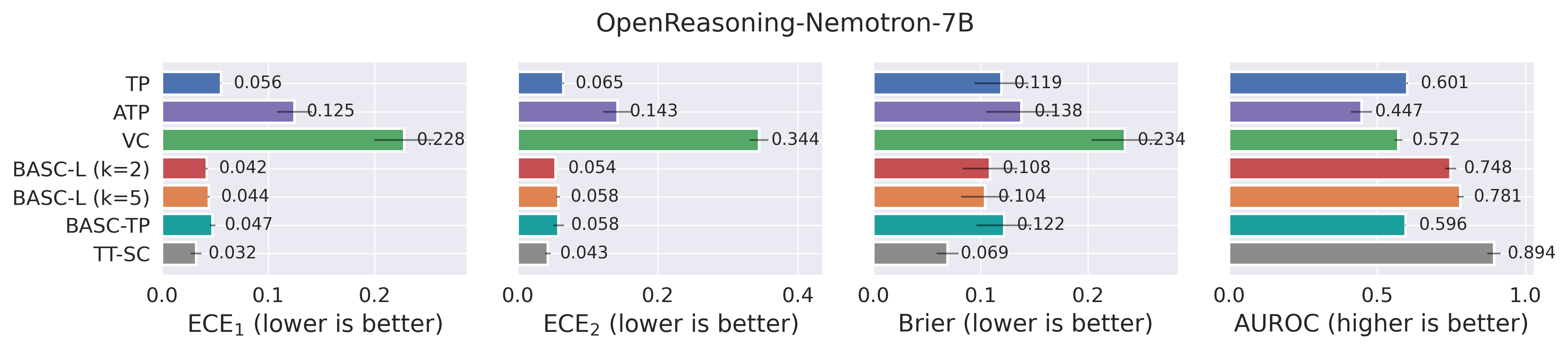}
	\caption{
    Main results for non-Qwen models.
    }
	\label{fig:res_per_model_non_qwen}
\end{figure}

\begin{figure}[!t]
	\centering
	\includegraphics[width=\textwidth]{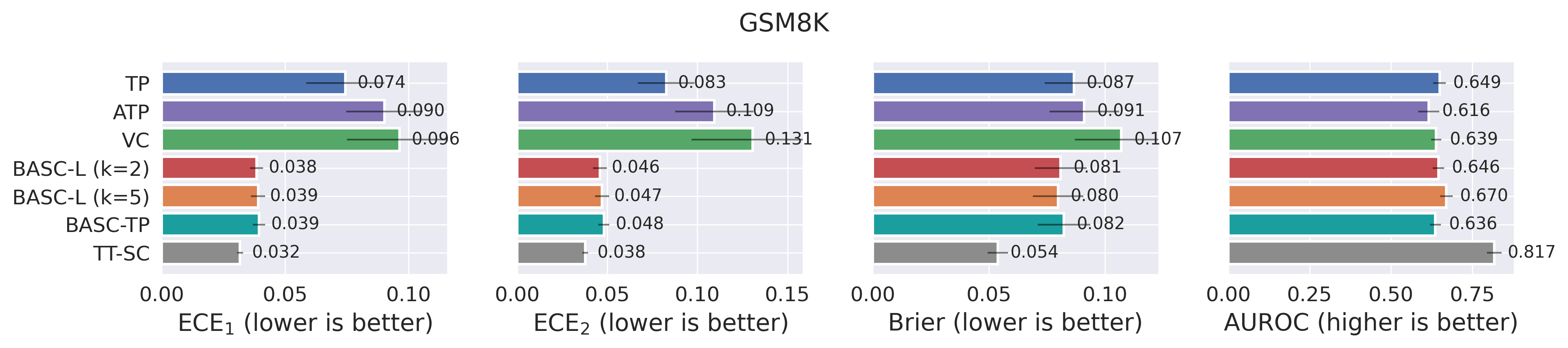}
    \includegraphics[width=\textwidth]{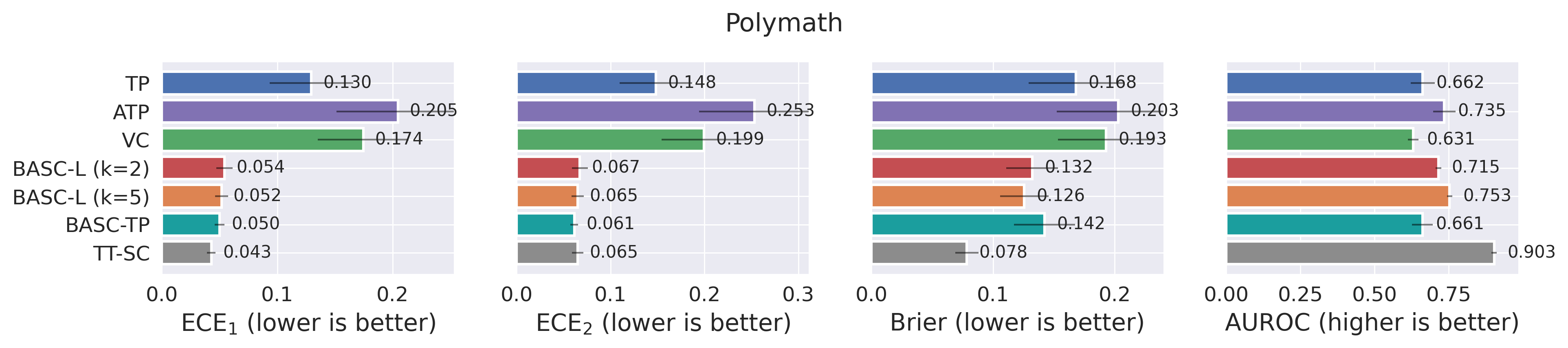}
    \includegraphics[width=\textwidth]{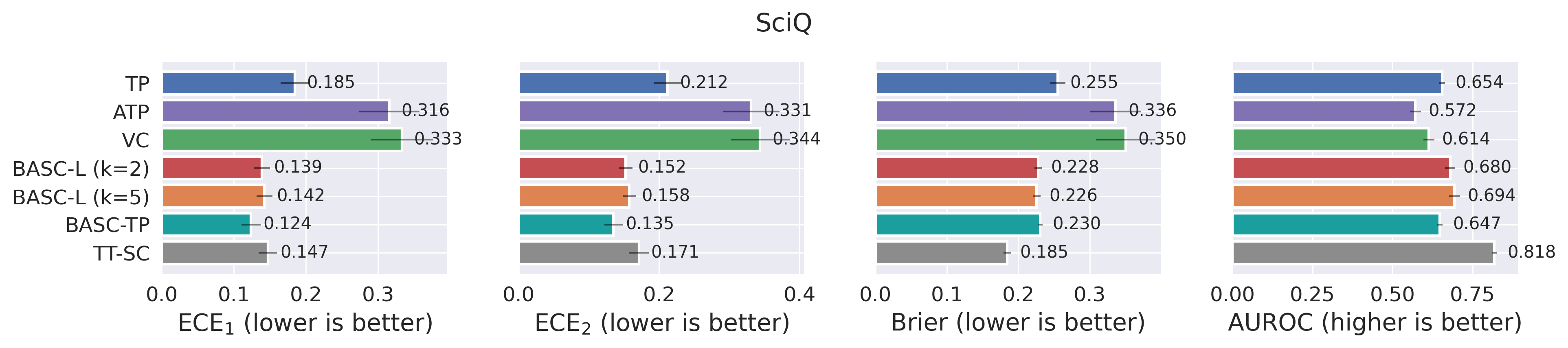}
	\includegraphics[width=\textwidth]{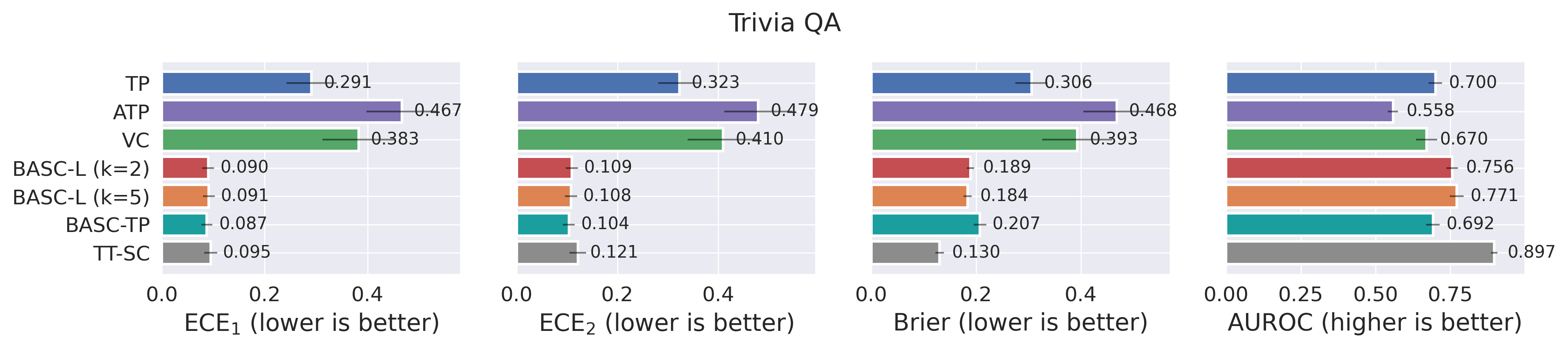}
    \includegraphics[width=\textwidth]{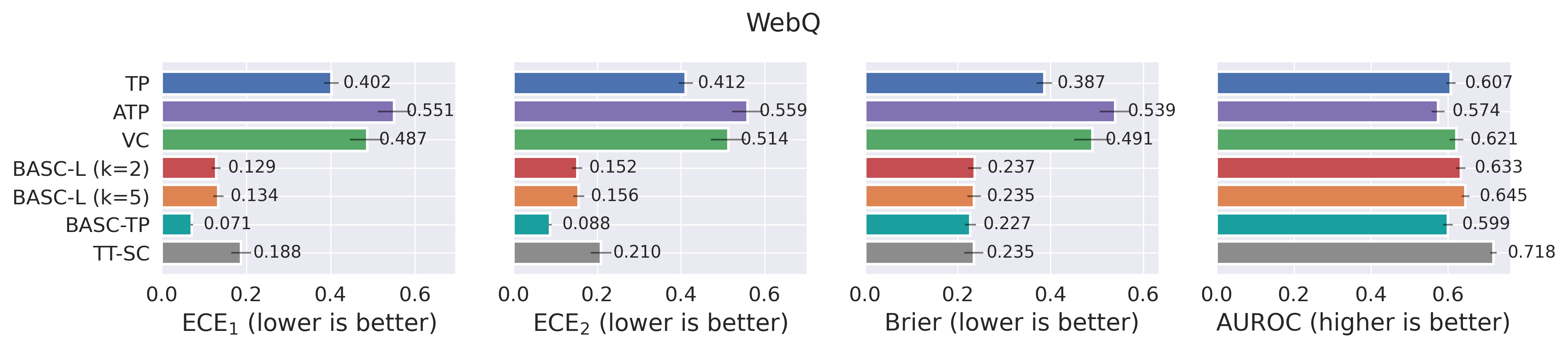}
	\caption{
    Main results for each dataset.
    }
	\label{fig:res_per_dataset}
\end{figure}

\begin{figure}[!t]
	\centering
	\includegraphics[width=\textwidth]{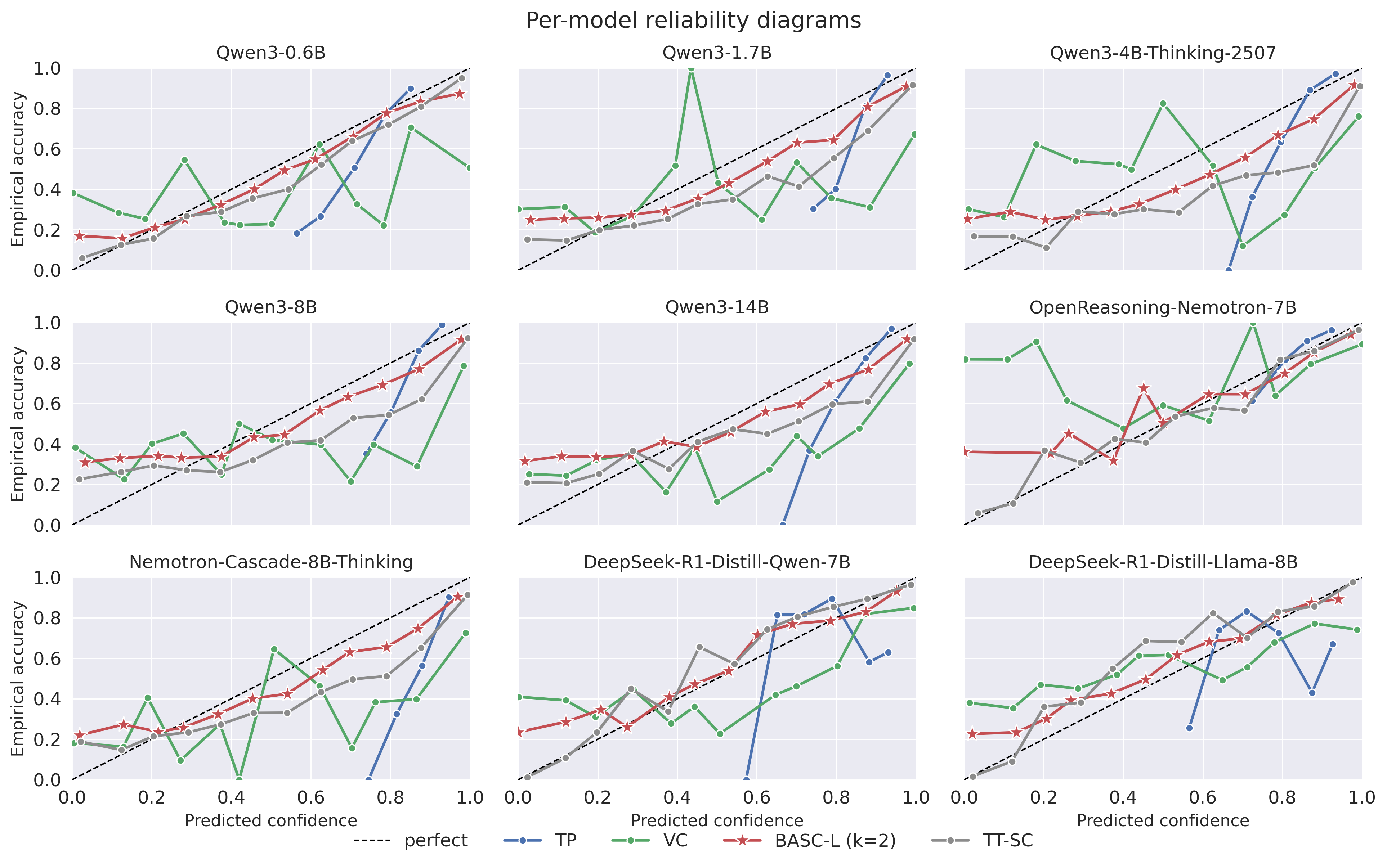}
	\caption{
    Reliability diagrams per model.
    }
	\label{fig:rel_diag_per_model}
\end{figure}

\begin{figure}[!t]
	\centering
	\includegraphics[width=\textwidth]{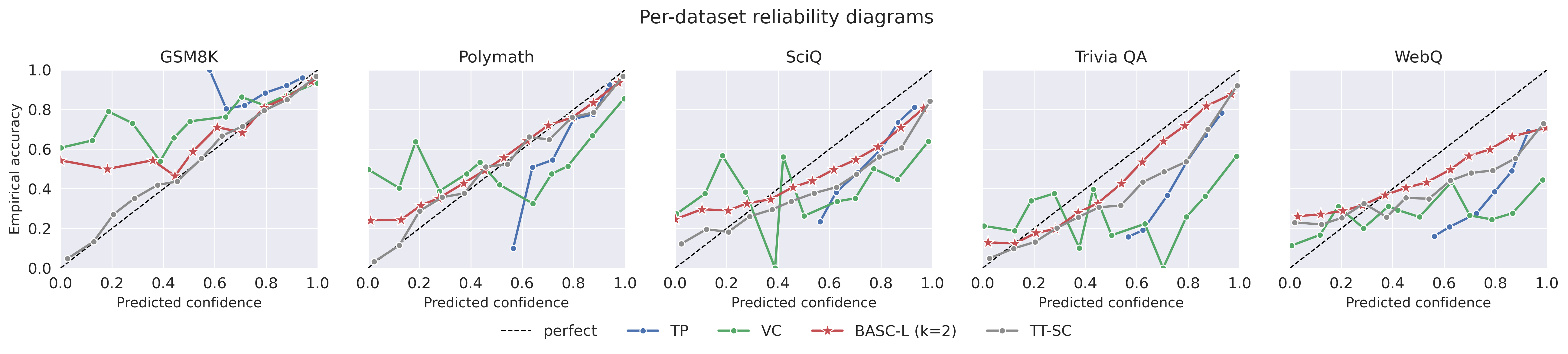}
	\caption{
    Reliability diagrams per dataset.
    }
	\label{fig:rel_diag_per_dataset}
\end{figure}

Accuracy for all model/dataset combinations is reported in Table \ref{tab:all_acc}.  Main results for Qwen models are reported in Figure \ref{fig:res_per_model_qwen}; for non-Qwen models are reported in Figure \ref{fig:res_per_model_non_qwen}; and for each dataset are reported in Figure \ref{fig:res_per_dataset}; standard error measurements are across datasets (for models) and models (for datasets). 
Reliability diagrams are also presented for each model (Figure \ref{fig:rel_diag_per_model}) and dataset (Figure \ref{fig:rel_diag_per_dataset}).

\begin{table}[!t]
    \centering
    \begin{tabular}{lcccccc}
    \toprule
    Model & GSM8K & Polymath & SciQ & Trivia QA & WebQ & Avg. \\
    \midrule
    Qwen3-0.6B & 0.806 & 0.643 & 0.409 & 0.215 & 0.205 & 0.456 \\
    Qwen3-1.7B & 0.904 & 0.841 & 0.599 & 0.376 & 0.353 & 0.615 \\
    Qwen3-4B-Thinking-2507 & 0.949 & 0.921 & 0.664 & 0.544 & 0.417 & 0.699 \\
    Qwen3-8B & 0.950 & 0.924 & 0.665 & 0.640 & 0.444 & 0.725 \\
    Qwen3-14B & 0.947 & 0.921 & 0.693 & 0.695 & 0.460 & 0.743 \\
    OpenReasoning-Nemotron-7B & 0.897 & 0.821 & — & — & — & 0.859 \\
    Nemotron-Cascade-8B-Thinking & 0.929 & 0.891 & 0.668 & 0.528 & 0.407 & 0.685 \\
    DeepSeek-R1-Distill-Qwen-7B & 0.913 & 0.614 & — & — & — & 0.764 \\
    DeepSeek-R1-Distill-Llama-8B & 0.869 & 0.460 & — & — & — & 0.665 \\
    \bottomrule
    \end{tabular}
    \caption{Accuracy for all model/dataset combinations.  Each row is a model, and our 5 datasets are on the columns.}
    \label{tab:all_acc}
\end{table}

\clearpage
\subsection{Ablations}\label{app:ablations}

Given our unsupervised setting, we perform no model- or dataset-specific tuning of our BASC method.
All method and sampling hyperparameters are fixed across the 36 model/dataset combinations in our main paper experiments.
However, to understand the sensitivity to potential choices, here we ablate probe sampling temperature in $[0.5, 0.7, 0.9]$ and QA answer matching method. 
We also discuss the computational costs of potential answer matching methods.

\begin{figure}[!t]
	\centering
	\includegraphics[width=\textwidth]{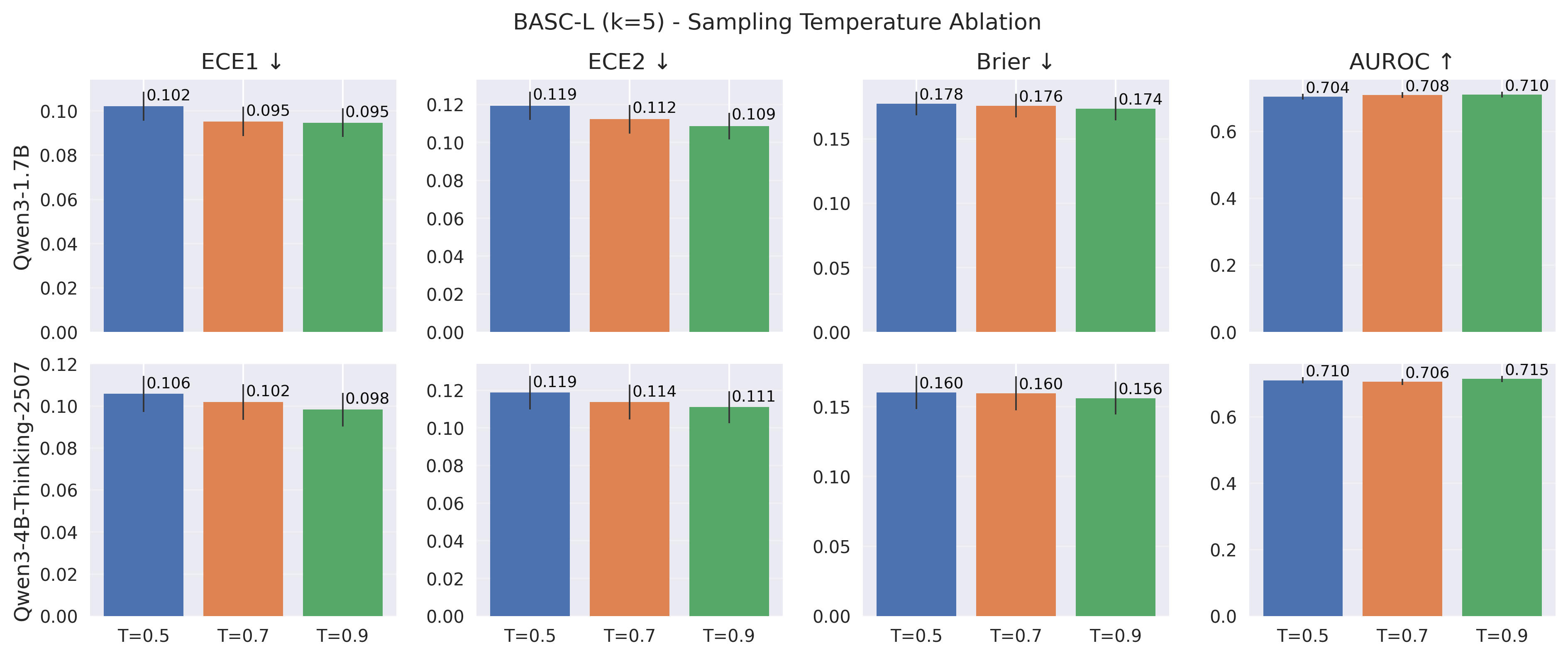}
	\caption{
    Ablating temperature for sampling probe responses for training our BASC recalibrator.
    }
	\label{fig:temp_ablation}
\end{figure}

\paragraph{Ablating probe sampling temperature.}
We ablate the probe sampling temperature over $T\in\{0.5,0.7,0.9\}$ for BASC-L ($k=5$) with two models in Figure~\ref{fig:temp_ablation}.
On the two evaluated models, performance is largely insensitive to this choice across calibration, Brier score, and AUROC, although higher temperatures show a slight tendency to improve performance.
Given our unsupervised setting, we retain our initial choice of $T=0.7$ as our fixed default rather than tuning it based on these results.

\begin{figure}[!t]
	\centering
	\includegraphics[width=\textwidth]{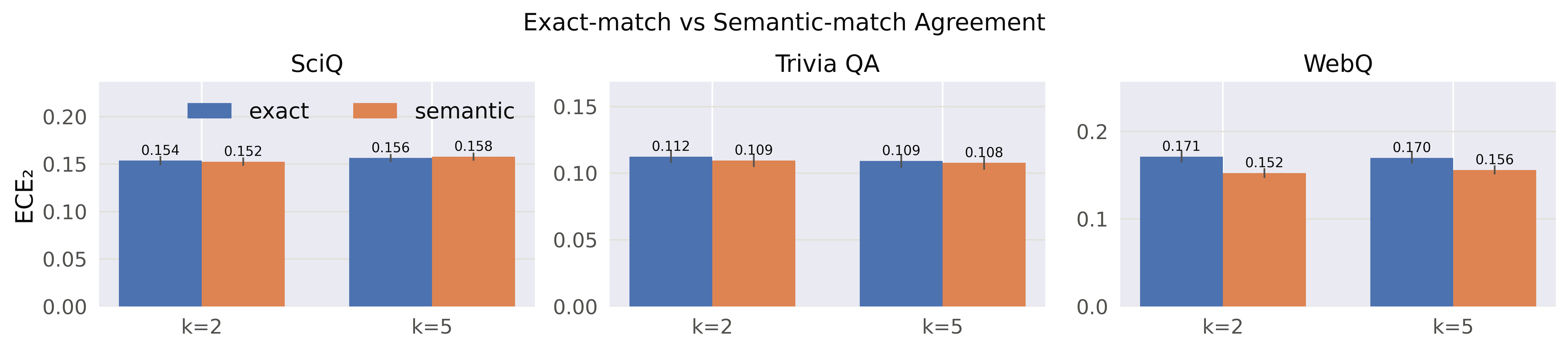}
	\caption{
    Ablating answer matching method.
    }
	\label{fig:matching_ablation}
\end{figure}

\paragraph{Ablating answer matching method for QA.}
We compare our semantic agreement criterion to normalized exact string matching on the QA tasks in Figure~\ref{fig:matching_ablation}.
The two produce similar calibration, with semantic matching providing a modest improvement on WebQ but no significant overall difference, indicating that BASC is not sensitive to the particular matching procedure.

\paragraph{Cost of semantic matching.} On the QA datasets, BASC measures agreement with bidirectional entailment, which costs one NLI forward pass in each direction per anchor–probe pair. 
This adds little computational cost compared to generating the responses. 
Each pair is two passes of DeBERTa-v3-base (approximately 86M non-embedding parameters) over about 55 tokens (question plus both answers), roughly 19 GFLOPs in total. 
We measured about 0.5 ms per pair on one A100 in fp16. 
A reasoning-model decode instead costs about $2PL$ FLOPs for $P$ parameters and $L$ generated tokens, and our responses average 530–1,740 tokens. 
Table~\ref{tab:nli_cost} shows that one agreement check costs 0.06–3\% of one decode. 
Because BASC compares each probe only with its anchor, the number of pairs grows in step with the number of decodes: BASC-L with $k=2$ uses 400 pairs alongside 800 decodes, adding at most 1.5\% to the calibration budget. 
Decoding generates tokens one at a time and is limited by memory bandwidth, whereas the NLI check is a single pass over a short input, so the wall-clock overhead is smaller still. 
Exact-match agreement on the math datasets requires no model at all. We therefore report calibration cost in full-response decodes only.

\begin{table}[!t]
    \centering
    \begin{tabular}{lrrrr}
    \toprule
    Model & Mean tokens & TFLOPs / decode & NLI pair / decode & Overhead \\
    \midrule
    Qwen3-0.6B                   &   534 &  0.64 & 2.97\% & 1.49\% \\
    Qwen3-1.7B                   & 1,181 &  4.02 & 0.47\% & 0.24\% \\
    Qwen3-4B-Thinking-2507       & 1,610 & 12.88 & 0.15\% & 0.07\% \\
    Qwen3-8B                     & 1,741 & 27.86 & 0.07\% & 0.03\% \\
    Qwen3-14B                    & 1,121 & 31.38 & 0.06\% & 0.03\% \\
    \bottomrule
    \end{tabular}
    \caption{
    Cost of semantic agreement relative to decoding on the QA datasets.
    Mean tokens: average length of the deployed response over SciQ, TriviaQA, and WebQ.
    Decode cost is approximated as $2PL$ FLOPs for $P$ nominal parameters and $L$ generated tokens (ignoring the prompt and attention).
    An NLI pair is two DeBERTa-v3-base forward passes over $\approx$55 tokens ($\approx$19 GFLOPs; $\approx$0.5\,ms on one A100 in fp16).
    Overhead: extra cost from 400 NLI pairs relative to its 800 decodes when running BASC-L ($k=2$).
    }
    \label{tab:nli_cost}
\end{table}

\clearpage
\subsection{Distribution Shift}\label{app:ood_exp}

\begin{figure}[!t]
	\centering
	\includegraphics[width=\textwidth]{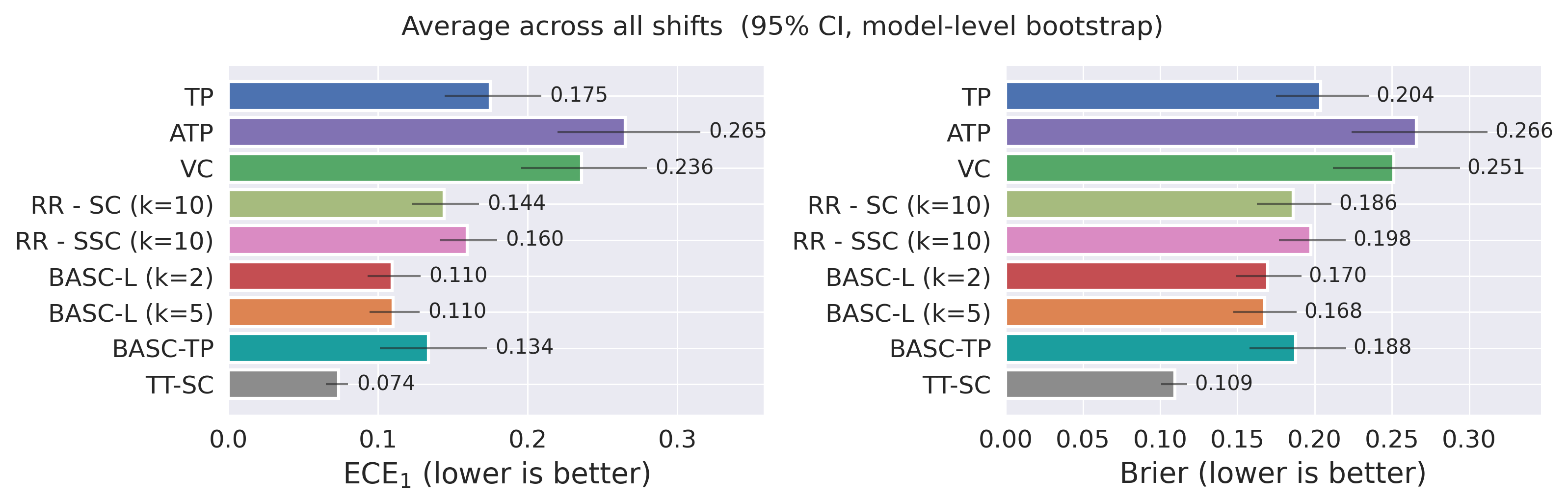}
	\caption{
    OOD results averaged across all shifts, with 95\% confidence intervals.
    }
	\label{fig:ood_avg_all_shifts}
\end{figure}

\begin{figure}[!t]
	\centering
	\includegraphics[width=\textwidth]{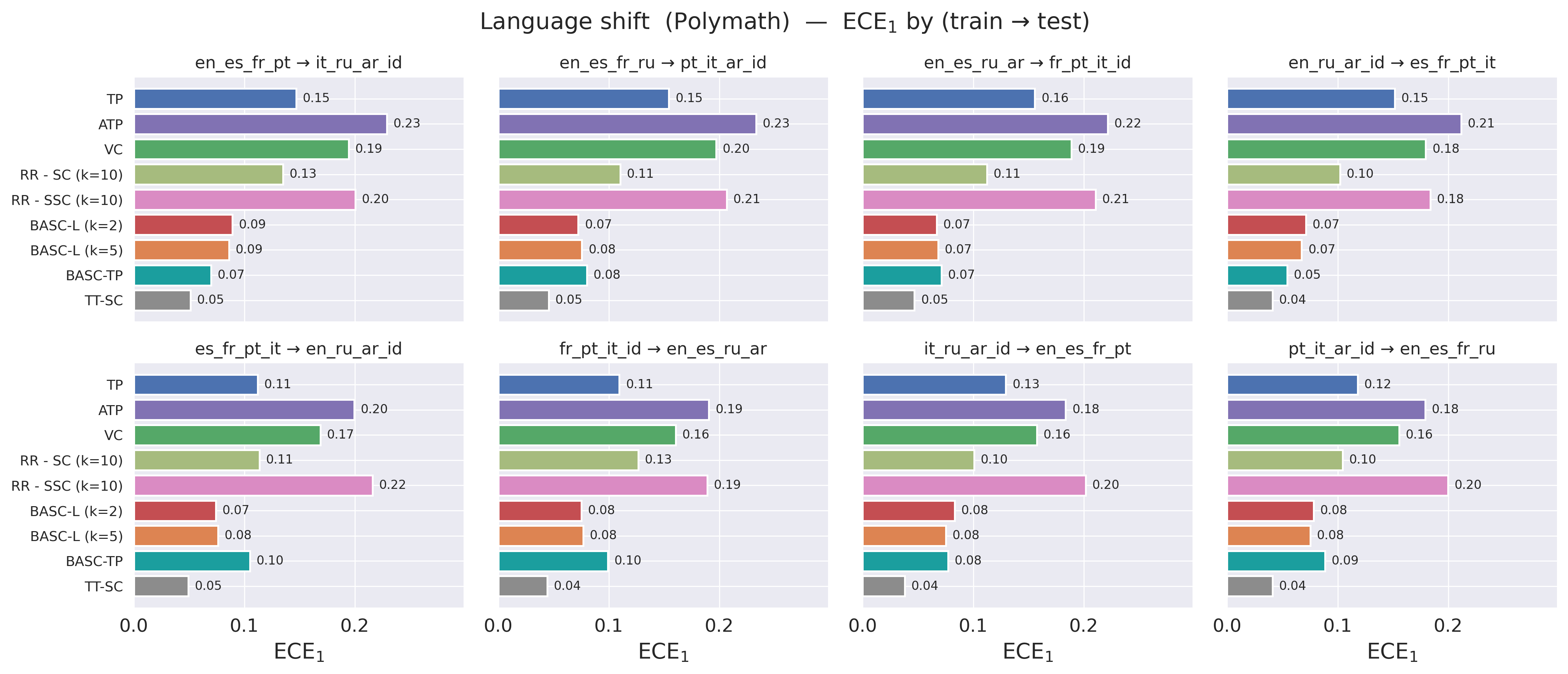}
    \includegraphics[width=\textwidth]{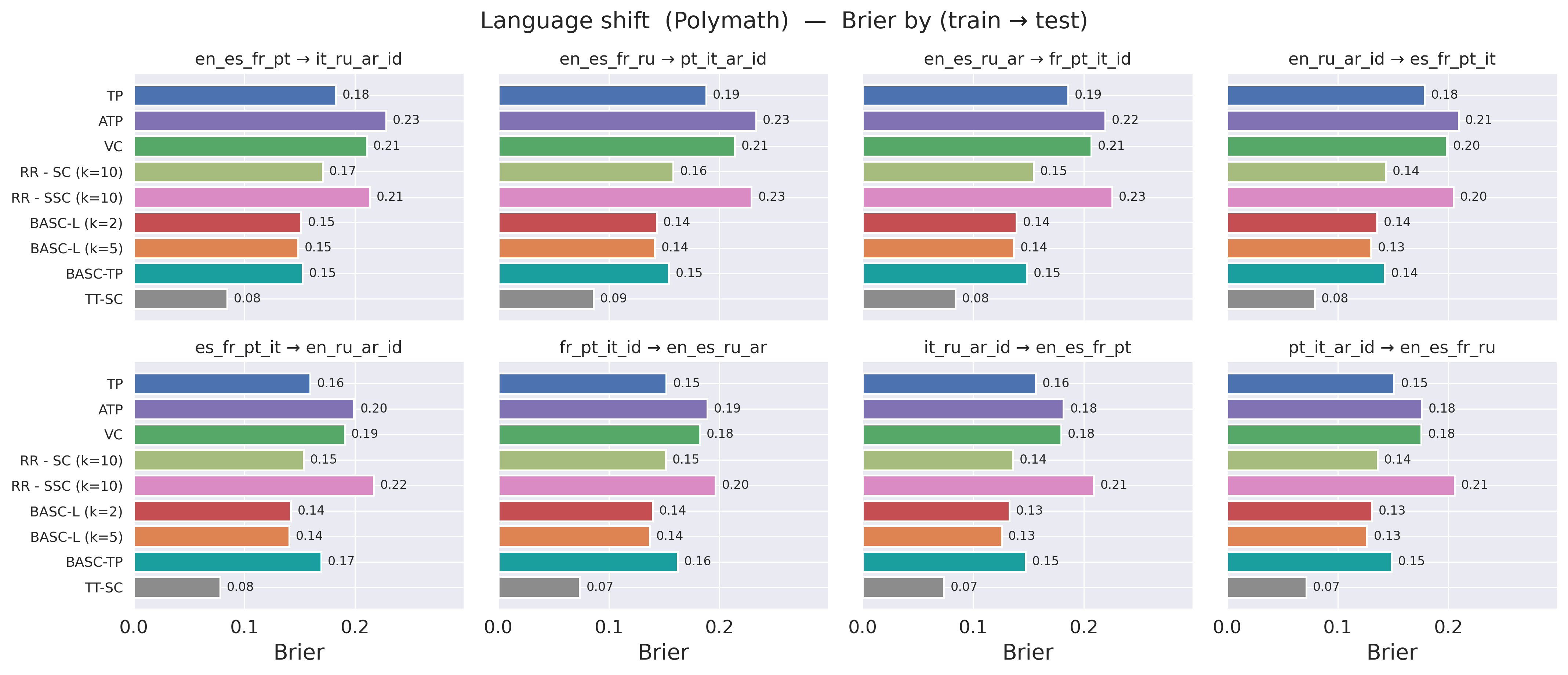}
	\caption{
    OOD results per language shift.
    }
	\label{fig:ood_language_shifts}
\end{figure}

\begin{figure}[!t]
	\centering
	\includegraphics[width=0.5\textwidth]{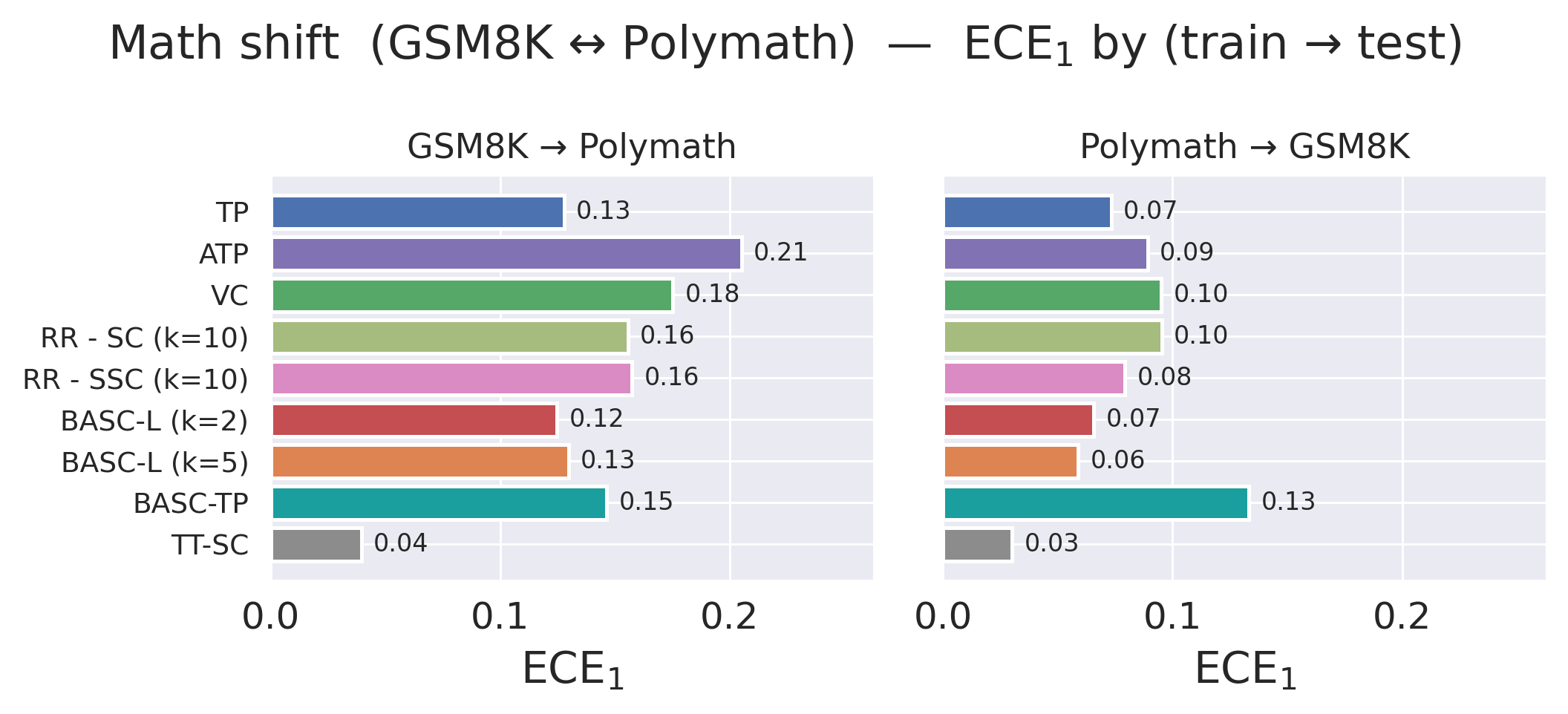}
    \includegraphics[width=0.5\textwidth]{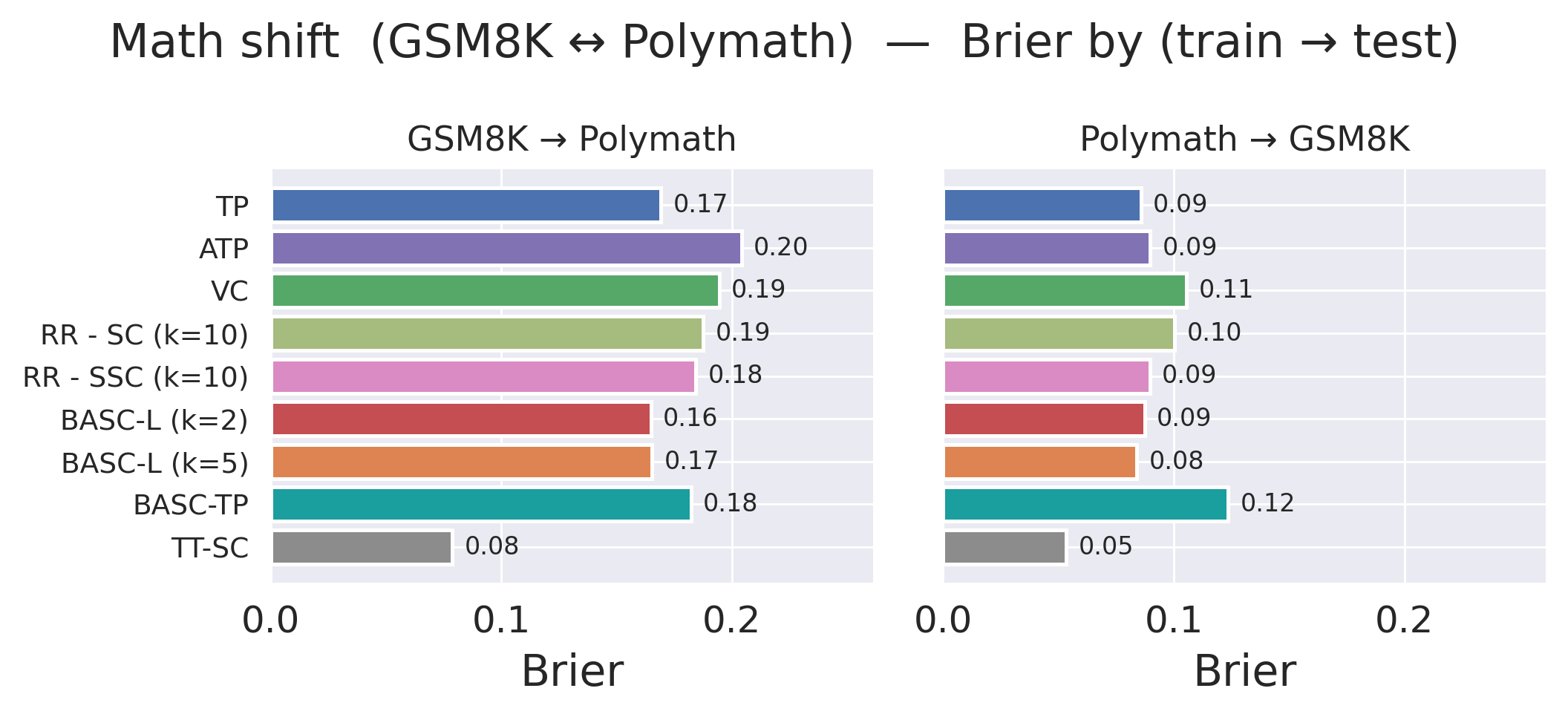}
	\caption{
    OOD results per math shift.
    }
	\label{fig:ood_math_shifts}
\end{figure}

\begin{figure}[!t]
	\centering
	\includegraphics[width=0.7\textwidth]{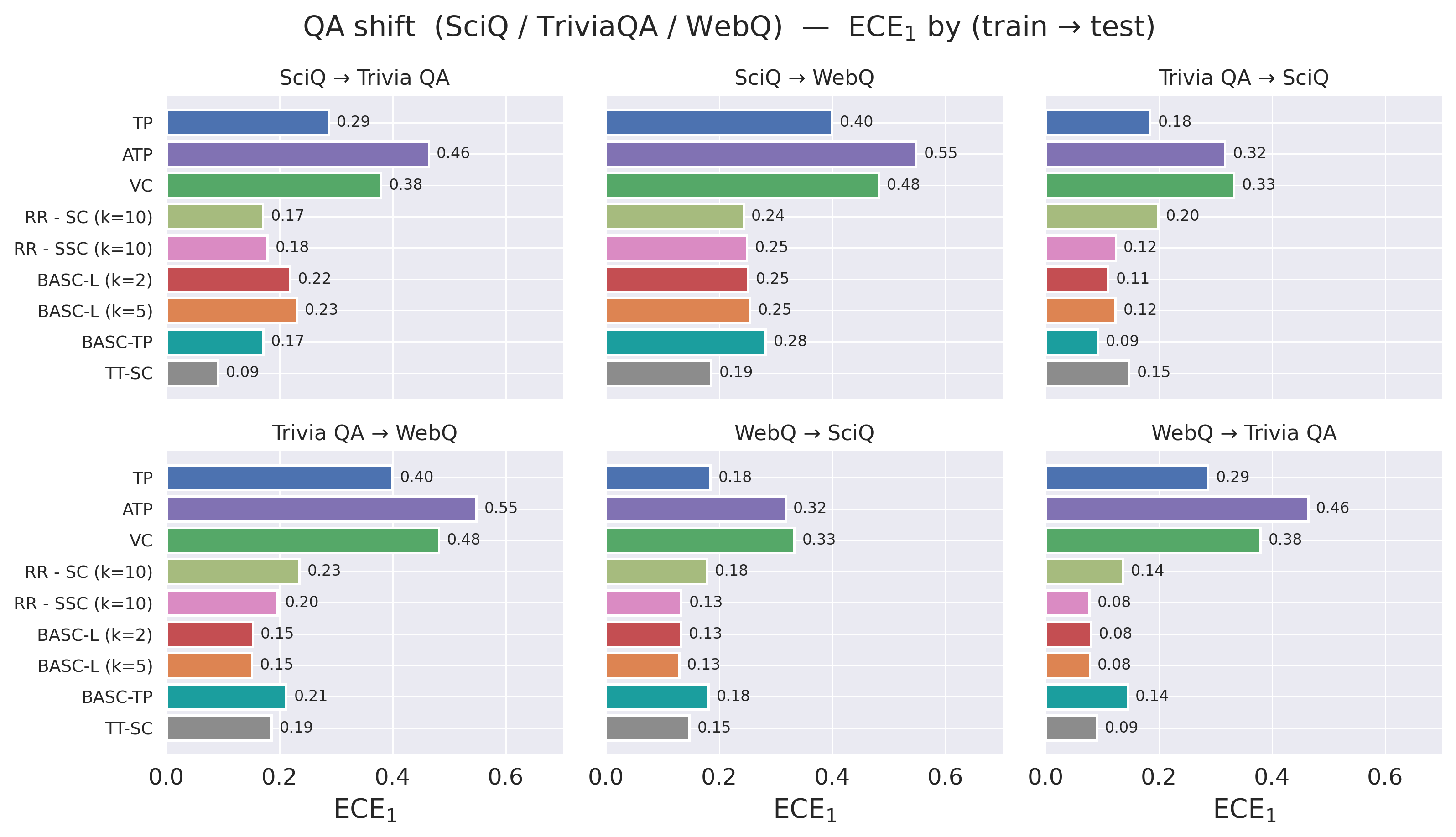}
    \includegraphics[width=0.7\textwidth]{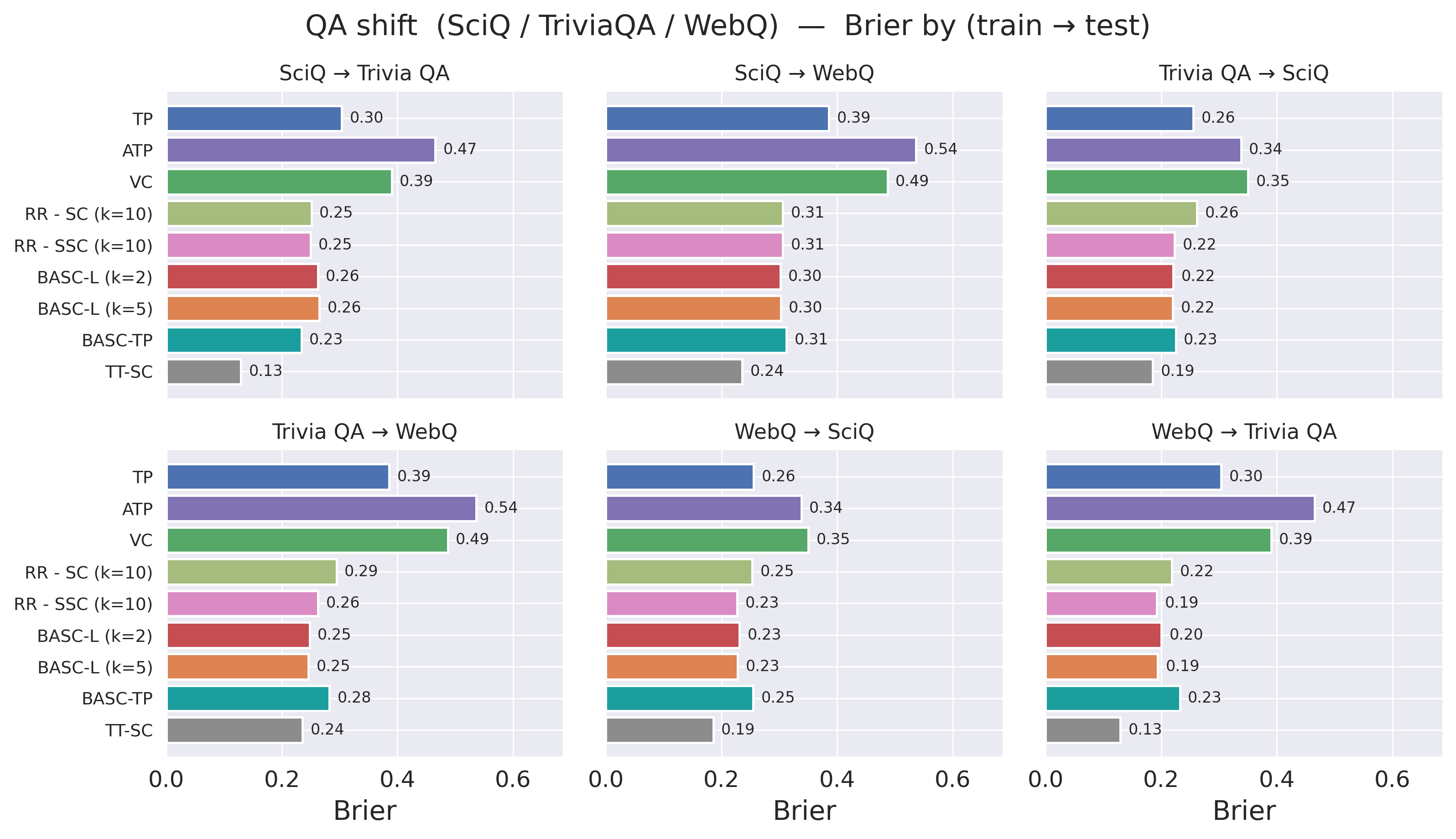}
	\caption{
    OOD results per QA shift.
    }
	\label{fig:ood_qa_shifts}
\end{figure}

Average results across all distribution shifts are presented in Figure \ref{fig:ood_avg_all_shifts}, including 95\% confidence intervals.
Results for each individual shift are presented in Figures \ref{fig:ood_language_shifts} (Language), \ref{fig:ood_math_shifts} (Math), and \ref{fig:ood_qa_shifts} (QA).

\clearpage
\subsection{Selective Prediction}

\begin{figure}[!t]
	\centering
	\includegraphics[width=\textwidth]{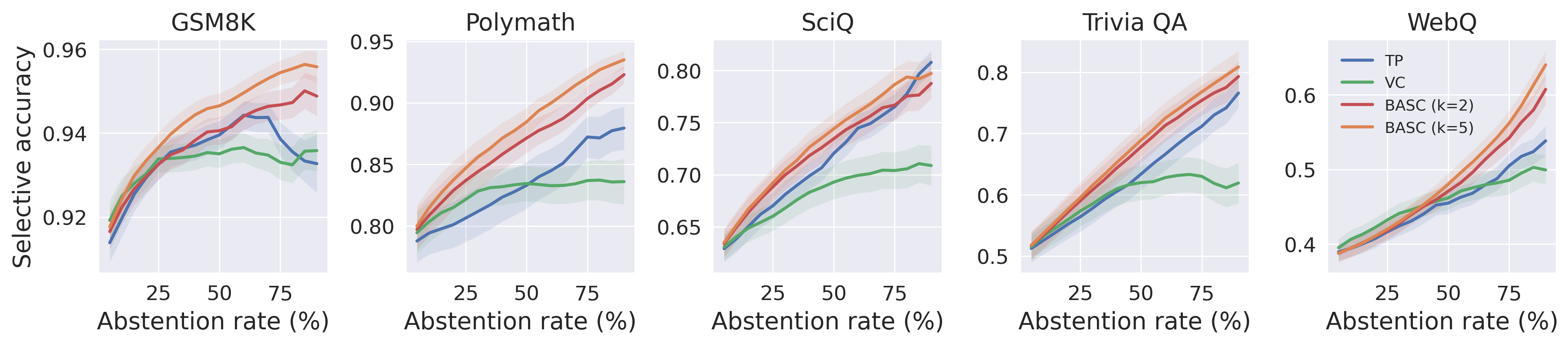}
	\caption{
    Selection results per dataset.
    }
	\label{fig:sel_by_dataset}
\end{figure}

\begin{figure}[!t]
	\centering
	\includegraphics[width=\textwidth]{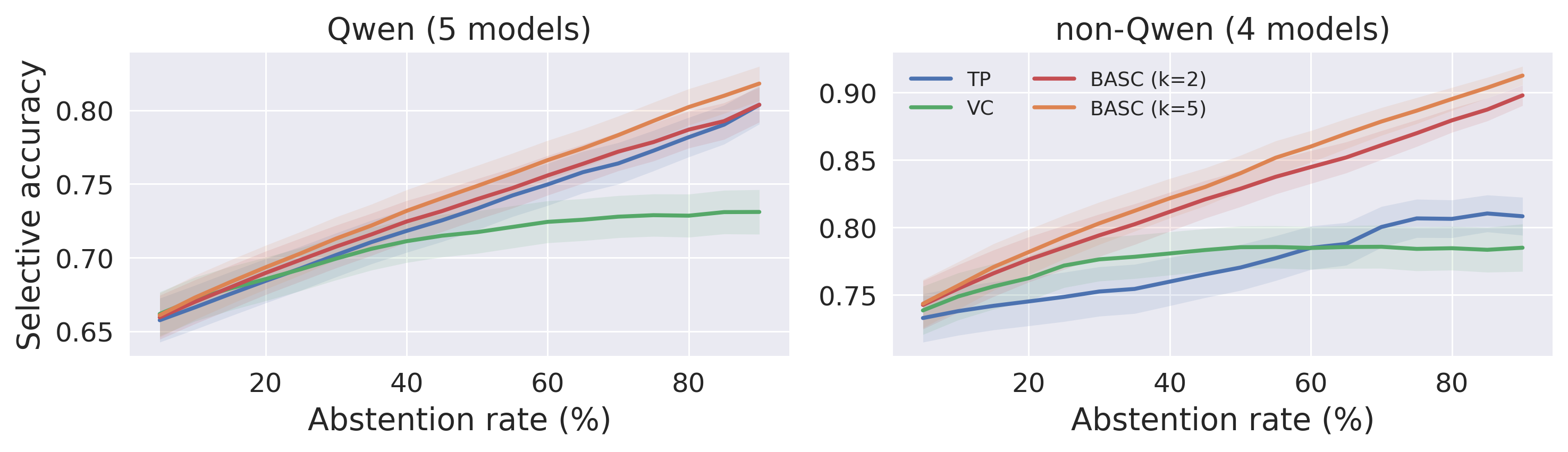}
	\caption{
    Selection results per model family.
    }
	\label{fig:sel_by_model}
\end{figure}

Selective prediction results are presented for each dataset (Figure \ref{fig:sel_by_dataset}) and model family (Figure \ref{fig:sel_by_model}).

\end{document}